\documentclass[lettersize,journal]{IEEEtran}
\usepackage{amsmath,amsfonts}
\usepackage{algorithmic}
\usepackage{algorithm}
\usepackage{array}
\usepackage[caption=false,font=normalsize,labelfont=sf,textfont=sf]{subfig}
\usepackage{textcomp}
\usepackage{stfloats}
\usepackage{url}
\usepackage{verbatim}
\usepackage{graphicx}
\usepackage{cite}
\usepackage{amssymb}
\usepackage{color,soul}
\usepackage{amsthm}
\newtheorem{thm}{Theorem}[section]

\newtheorem{proposition}[thm]{Proposition}

\DeclareMathOperator*{\argmax}{arg\,max}

\begin{document}

\title{Understanding Uncertainty-based Active Learning Under Model Mismatch} 


\author{
Amir Hossein Rahmati, Mingzhou Fan, Ruida Zhou, Nathan M. Urban, Byung-Jun Yoon,~\IEEEmembership{Senior Member,~IEEE,}, Xiaoning Qian,~\IEEEmembership{Senior Member,~IEEE,} 

\thanks{Amir Hossein Rahmati, Mingzhou Fan, Byung-Jun Yoon, and Xiaoning Qian are with the Department of Electrical and Computer Engineering, Texas A\&M University, College Station, TX 77843 USA (e-mail: amir\_hossein\_rahmati@tamu.edu; mzfan@tamu.edu; bjyoon@tamu.edu; xqian@tamu.edu)}
\thanks{Ruida Zhou is with the Department of Electrical and Computer Engineering, University of California, Los Angeles, CA 90095 (e-mail: ruida@g.ucla.edu)}
\thanks{Nathan M. Urban, Byung-Jun Yoon, and Xiaoning Qian are with Brookhaven National Laboratory, Upton, NY 11973 (e-mail: nurban@nbl.gov; byoon@bnl.gov; xqian1@bnl.gov).}
}



\IEEEpubid{This work has been submitted to the IEEE for possible publication. Copyright may be transferred without notice, after which this version may no longer be accessible.}

\maketitle

\begin{abstract}
Instead of randomly acquiring training data points, Uncertainty-based Active Learning~(UAL) operates by querying the label(s) of pivotal samples from an unlabeled pool selected based on the prediction uncertainty, thereby aiming at minimizing the labeling cost for model training.  
The efficacy of UAL critically depends on the model capacity as well as the adopted uncertainty-based acquisition function.  
Within the context of this study, our analytical focus is directed toward comprehending how the capacity of the machine learning model may affect UAL efficacy. 
Through theoretical analysis, comprehensive simulations, and empirical studies, we conclusively demonstrate that UAL can lead to worse performance in comparison with random sampling when the machine learning model class has low capacity and is unable to cover the underlying ground truth. In such situations, adopting acquisition functions that directly target estimating the prediction performance may be beneficial for improving the performance of UAL.
\end{abstract}


\section{Introduction}\label{Introduction}
\IEEEPARstart{W}{ith} the advent of novel profiling and database technologies, the ever-increasing volume of available data gives rise to extensive unlabeled datasets that may help further advance the AI and machine learning~(AI/ML) development~\cite{WANG202012}. However, the associated cost of labeling is exorbitantly prohibitive \cite{Wang_2017, 8862913, gal2017deep, 8622459}, to which Active Learning (AL) can be a potential solution.

AL aims to reduce the quantity of labeled training data while achieving the desired prediction performance in ML~\cite{settles2009active}. Based on a selected acquisition function, AL iteratively queries the labels of the most informative samples,  hoping to learn with better sample efficiency utilizing much fewer samples than what is available in the initial unlabeled dataset~\cite{ren2021survey}. 
This iterative AL procedure persists until either the model achieves the desired prediction performance or the designated labeling budget is exhausted. 
These methods broadly can be divided into \emph{membership query synthesis}\cite{10.1023/A:1022821128753, settles2009active, ren2021survey}, \emph{stream-based selective sampling} \cite{DAGAN1995150, ren2021survey, settles2009active}, and \emph{pool-based active learning} \cite{LEWIS1994148, ren2021survey, settles2009active} 
based on the problem setup.
There are two main categories of sample selection strategies considered in pool-based active learning methods: \emph{Uncertainty-based AL}~(UAL)~\cite{lewis1995sequential,settles2009active, 6889457, 10.1007/11871842_40, 10.1162/153244302760185243, Tr2005CombiningAA} and \emph{Diversity-based AL}~(DAL)~\cite{settles2008curious,settles2008analysis,ren2021survey,settles2009active, sener2017active, geifman2017deep}. 
Some studies also focused on using both categories leading to hybrid methods~\cite{wu2022entropybased, ash2020deep}. For UAL, samples are selected based on the model's prediction confidence/uncertainty reflecting their significance in improving the model's performance. 
Variance-based or information-theoretic methods (estimated by entropy or mutual information), such as Bayesian Active Learning by Disagreement~(BALD)~\cite{houlsby2011bayesian}, are considered under this category. 
DAL strives to identify the subset encapsulating the underlying data distribution, including core-set methods~\cite{sener2017active}. 
This study centers its attention on UAL, which identifies and queries labels for the most uncertain samples, implying their importance for the current model training~\cite{settles2009active}. 

\begin{figure}[t]\label{fig: gp_mot}

   \centerline{\includegraphics[width=1.02\linewidth]{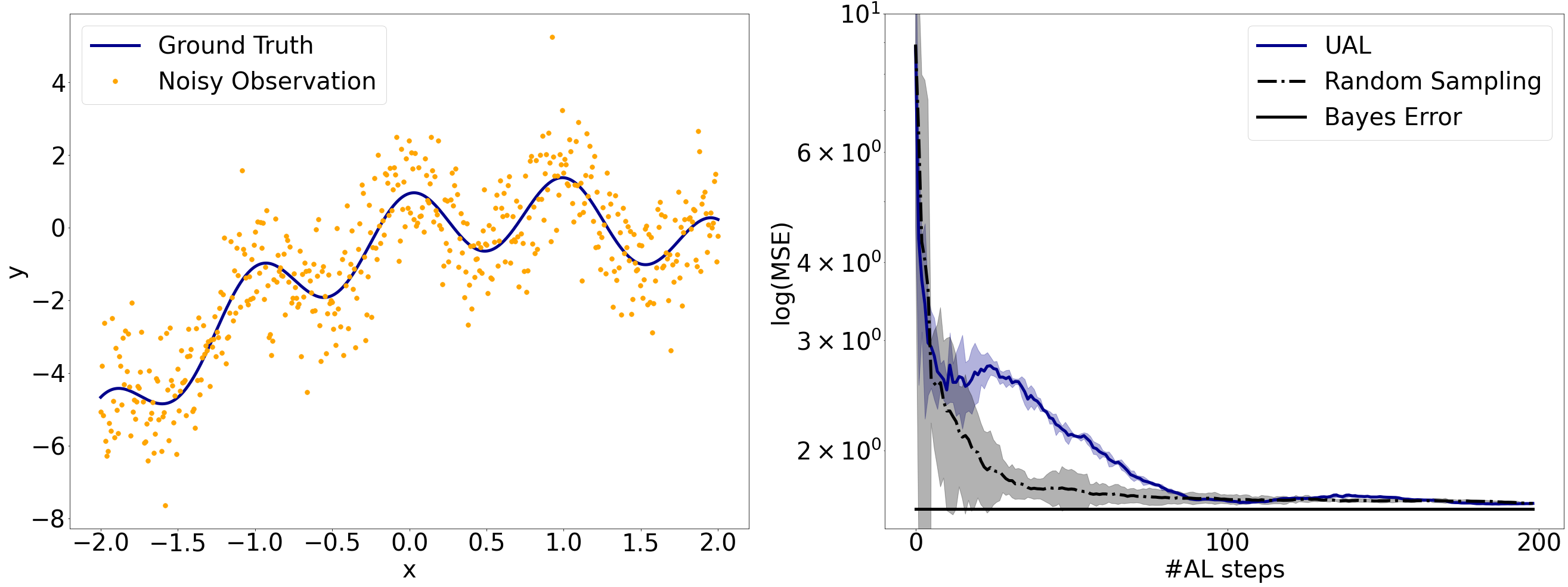}}
    \caption{Motivating example, where the ground truth is more complex than what can be captured by the prediction model class. In this example, the noisy data follows the ground truth function that is the summation of a quadratic function, and a cosine function ( $y=f(x)+\epsilon$ with $\epsilon \sim N(0,1)$, where $f(x) = \langle\mathbf{x},\mathbf{w}\rangle + cos(2\pi x)$, and $\mathbf{x} = [1,x,x^2]$). The left plot shows the ground truth target function and its corresponding noisy observed data. 
    The right plot shows the performance comparison between UAL and random sampling, 
    to learn the quadratic predictor based on Bayesian polynomial regression.\\
    }
    \label{fig: Fig(motHT)}
\end{figure}

Despite the popularity of UAL methods, some previous works have demonstrated situations where they may not outperform random sampling~\cite{hacohen2022active,munjal2022robust,saifullah2023analyzing, sinha2019variational}, which motivates this study. To better illustrate the need to understand 
the behaviour of UAL performance,
we first present such an example 
where 
we compare the regression performance of UAL and random sampling with a quadratic model on data generated from a more complex ground truth function in Figure~\ref{fig: Fig(motHT)}. 
Instead of converging faster than random sampling as typically expected, 
UAL fails to effectively select the most informative samples to better guide the model training with the prediction performance of learned regressors worse than the ones trained with random sampling.

\IEEEpubidadjcol
These observations have triggered our curiosity to investigate the potential effects of the model capacity on the efficiency of UAL. We seek to find the necessary 
settings from the model's perspective, under which UAL can outperform random sampling. In particular, we establish that given a model of at least adequate complexity covering the underlying ground truth target function to learn, UAL is able to have superior performance than random sampling. 
Relying on this establishment, we recommend potential remedies to overcome UAL's pitfalls under model-mismatch scenarios and empirically showcase their effectiveness. In summary our contributions can be summarised as follows:

\begin{itemize}
\item We investigate the UAL's efficiency from the predictive model class perspective in regression task
\item We establish the necessary conditions under which UAL has the potential to perform efficiently both through theoretical analysis and empirical evaluations.
\item Built upon our analytical results, we recommend potential remedies for UAL's drawbacks under model-mismatch scenarios.
\end{itemize}


The subsequent sections are organized as follows: Section \ref{sec:Problem Settings} provides the necessary background, definitions, and UAL settings. 
In Section \ref{UAL efficacy}, 
we analyze the UAL efficacy under different model settings. In accordance with results in Section~\ref{UAL efficacy}, in Section~\ref{remedy} we investigate potential strategies to remedy UAL's drawbacks in model mismatch scenarios. 
We present experimental results supporting the findings of the analysis in Section \ref{experiments}. We then conclude the study 
in Section \ref{Conclusion}.

\section{Problem settings}\label{sec:Problem Settings}

In this work, we focus on the performance of pool-based UAL for regression under the metric of Mean Square Error~(MSE) for illustrative theoretical analysis. 
The regression problem is to assign $\mathbf{x} \in \mathcal{X}\subset\mathbb{R}^d$ to its corresponding output $y$ with the assumption that the input random vector $\mathbf{X}$ and output random variable $Y$ are jointly distributed following $P(\mathbf{X}, Y)$. 
The common setting is that, at an arbitrary $\mathbf{X}=\mathbf{x}$, the output is ${y} = f(\mathbf{x}) + \epsilon$, where $f:\mathcal{X} \rightarrow \mathbb{R}$ is the underlying ground truth function and $\epsilon \sim N(0,\sigma^2)$ is the white noise residual error determining the conditional distribution $P(Y=y|\mathbf{X}=\mathbf{x})$. 
Given an observed set of data points, $\mathbf{D}=\{(\mathbf{x}_i,{y}_i)\}_{i=1}^{n} \text{ with } (\mathbf{x}_i,{y}_i)\sim P(\mathbf{X},Y)$, 
$\hat{{y}} =\hat{f}_{{\theta}}(\mathbf{x}) + \epsilon$ 
serves as an estimator of $y$ 
in the regression task, where a regressor $\hat{f}_{\theta}:\mathcal{X}\rightarrow \mathbb{R}$, is a parameterized predictive model with $\theta$ being the model parameters.

To quantify the uncertainty of the learned model and further use it to guide UAL, we depend on Bayesian Learning in this study. In the following, we briefly introduce Bayesian Learning, Bayesian Active Learning, and the learning objective.

\subsubsection{Bayesian Learning}

Bayesian Learning aims to maintain the posterior distribution of the parameters $\theta$, $P^*(\theta|\mathbf{D}) \propto P(\theta)P(\mathbf{D}|\theta)$, based on the Bayes rule with the observed dataset $\mathbf{D}$, 
where $P(\theta)$ is the prior distribution reserving prior knowledge, and $P(\mathbf{D}|\theta)$ is the likelihood function representing the probability of generating the observed data with specific parameter $\theta$.
Via the parameter posterior, the predictive posterior 
at an arbitrary $\mathbf{x}$ can be calculated as follows:
\begin{equation}\label{eq:predctive_posterior}\footnotesize
\begin{aligned}
    \pi^*(\hat{y}|\mathbf{x}) &= \int P(\hat{y}|\theta,\mathbf{X}=\mathbf{x})P^*(\theta|\mathbf{D}) d\theta. 
\end{aligned}
\end{equation}
\subsubsection{Bayesian Active Learning (BAL)}\label{BAL}
\begin{figure}
    \centering
    \includegraphics[width=0.85\linewidth]{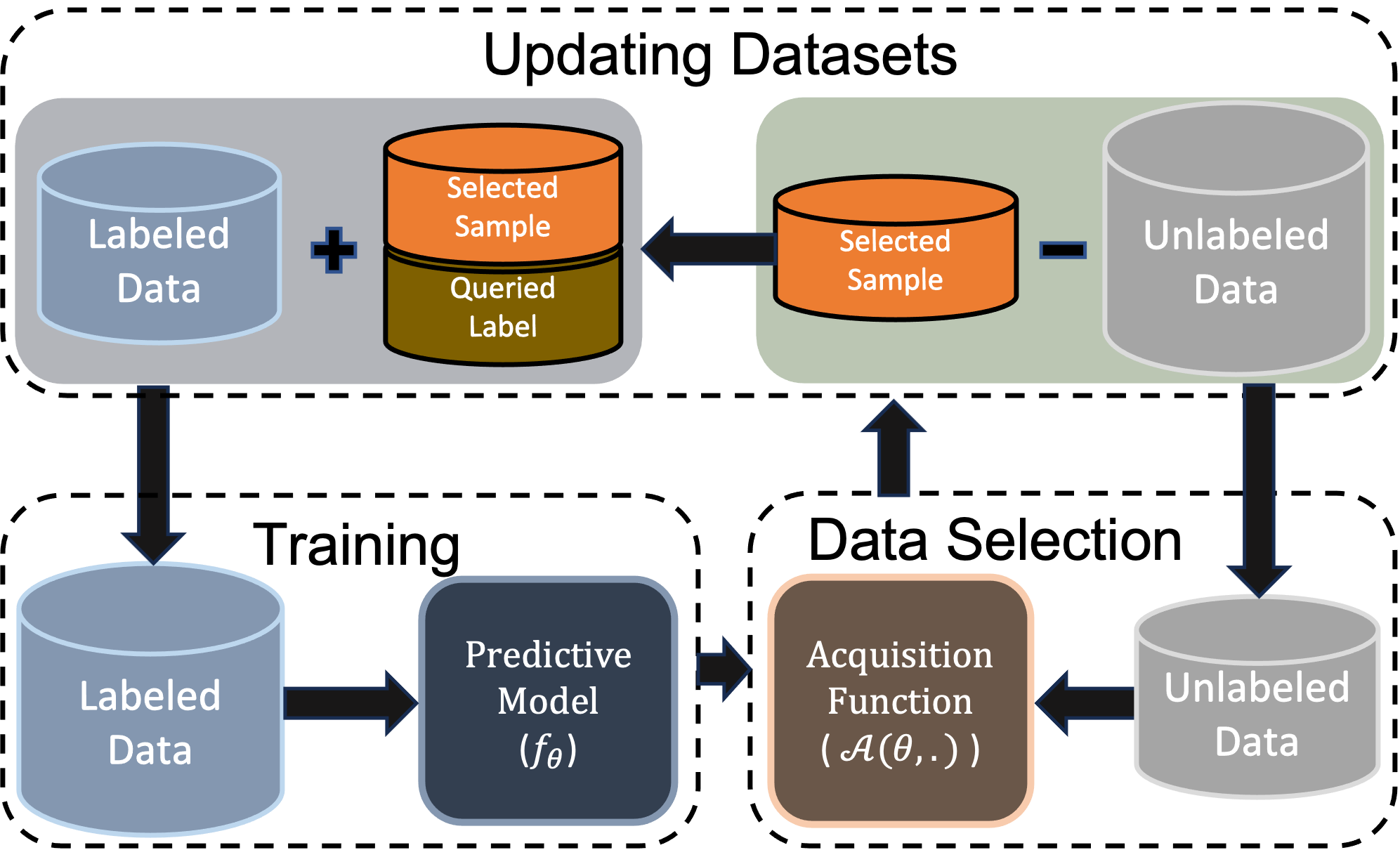}
    \caption{Schematic illustration of the UAL procedure.}
    \label{fig:UAL-diagram}
\end{figure}

In BAL, we consider the training dataset to be composed of labeled and unlabeled data, i.e. $\mathbf{D} = \mathbf{D_L} \bigcup \mathbf{D_U}$, where $\mathbf{D_L} = \{(\mathbf{x}_i,{y}_i)\}_{i=1}^{n_\mathbf{L}}$ is the labeled dataset and $\mathbf{D_U}=\{\mathbf{x}_j\}_{j=1}^{n_\mathbf{U}}$ is the unlabeled dataset, $n_\mathbf{L}$, and $n_\mathbf{U}$ are the number of labeled and unlabeled samples respectively. The BAL procedure aims at selecting the most informative sample in the unlabeled dataset by optimizing the acquisition function $\mathcal{A}$. The label of selected $\mathbf{x}^* \in \argmax_{\mathbf{x_U}\in \mathbf{D_U}} \mathcal{A}(\pi^*, \mathbf{x_U})$  
will then be queried and added to the labeled dataset, 
while $\mathbf{x}^*$ will be removed from the unlabeled dataset. 
This procedure ends when the performance is satisfactory or the budget is exhausted, 
$\mathbf{D_U} = \varnothing$. 
In the context of Bayesian UAL, equipped with learned posterior distributions, $\mathcal{A}$ is designed based on the prediction uncertainty that hopefully captures both uncertainties coming from data and model (the higher the score is, the higher the uncertainty is). With the same procedure as BAL, the most uncertain sample will be added iteratively to the labeled dataset at each UAL step. 
{Figure~\ref{fig:UAL-diagram} offers a sketch of this active learning process.}

\subsubsection{Learning Objective}
Our UAL performance analysis will be based on the MSE of prediction, $\hat{f}_\theta$, with respect to observed output, $Y$, i.e., $\text{MSE}_{\text{obs}}$, which can be written as: 
\[
    \begin{aligned}
        \text{MSE}_{\text{obs}} &= E_{P(Y|\mathbf{X})}[ E_{P^*(\theta|\mathbf{D_L})}  [(Y-\hat{f}_{\theta})^2]]  \\
        &= \int P(Y|\mathbf{X}) (\int P^*(\theta|\mathbf{D_L}) (Y-\hat{f}_{\theta})^2 d{\theta})dY, 
    \end{aligned}
\]
where $P$ denotes the ground truth distribution, and $P^*$ the derived posterior distribution given training data.

\section{UAL Efficacy Analysis} \label{UAL efficacy}

Following the problem setup in Section \ref{sec:Problem Settings}, we first decompose $\text{MSE}_{\text{obs}}$ for the latter analysis of UAL performance with different prediction model classes:  
\[
    \begin{aligned}
        \text{MSE}_{\text{obs}} &= E_{P(Y|\mathbf{X})}[ E_{P^*(\theta|\mathbf{D_L})}  [(Y- E_{P(Y|\mathbf{X})}[Y] \\
        &\quad + E_{P(Y|\mathbf{X})}[Y] - \hat{f}_{\theta})^2]] \\
        &=E_{P(Y|\mathbf{X})}[(Y- E_{P(Y|\mathbf{X})}[Y])^2]\\ 
        &\quad + E_{P(Y|\mathbf{X})}[ E_{P^*(\theta|\mathbf{D_L})} [(E_{P(Y|\mathbf{X})}[Y] -\hat{f}_{\theta})^2]].
    \end{aligned}
\]

In the above $\text{MSE}_{\text{obs}}$ decomposition, the first term comes from the observation noise, and the second term is related to the error arising from the predictive model settings. Since we aim to investigate UAL's performance from the model's perspective, our focus is on the analysis of the second term, i.e. $\text{MSE}$ of predicted output from the ground truth $f$, which can further decomposed as: 
\[
    \begin{aligned}
        \text{MSE}
        &= E_{P(Y|\mathbf{X})}\Big[ E_{P^*({\theta}|\mathbf{D_L})} [(\hat{f}_{\theta}-E_{P(Y|\mathbf{X})}[Y])^2]\Big] \\
        &= E_{P(Y|\mathbf{X})}\Big[ \underbrace{E_{P^*({\theta}|\mathbf{D_L})} [(\hat{f}_{\theta}-E_{P^*({\theta}|\mathbf{D_L})}[\hat{f}_{\theta}])^2]}_{\mbox{\textbf{Variance}}}\\ 
        &\quad + \underbrace{(E_{P^*(\theta|\mathbf{D_L})}[\hat{f}_{\theta}]-  E_{P(Y|\mathbf{X})}[Y])^2}_{\mbox{\textbf{Bias}}}\Big].
    \end{aligned}
\]

The above equation gives the well-known bias-variance decomposition. 
In the following, we will justify our choice of the acquisition function following similar assumptions in the Bernstein-von Mises theorem \cite{schervish2012theory}.

\begin{thm}[Bernstein-von Mises \cite{schervish2012theory}]\label{BVM}
Let $\mathbf{D_L} = \{(x_i,y_i)\}_{i=1}^{n_L} \overset{\mathrm{iid}}{\sim} P_{\theta_{*}}$, for $\theta_*, \theta \in \Theta \subset \mathbb{R}^p$. Let $\ell_n =  \log P(\mathbf{D_L} \mid \theta) $ and define $\zeta = \Sigma_n^{-\frac{1}{2}} (\theta - \hat{\theta}_n)$ where $\hat{\theta}_n = \argmax_{\theta} \log P(\mathbf{D_L} \mid \theta) $ (the MLE) and $\Sigma_n = [- \ell^{\prime\prime}({\hat{\theta}_n})]^{-1}$ is the inverse information matrix. Then under regularity conditions, and assuming prior puts positive mass around $\theta_*$, the posterior of $\zeta_n$ given $\mathbf{D_L}$ converges in probability uniformly on compact sets to $N(\mathbf{0}, I_p)$. That is for each compact subset $B\subseteq \Theta$, and each $\epsilon>0$,
\[
\lim_{n \rightarrow \infty} P(\sup_{\zeta \in B} \mid \pi_{\zeta|\mathbf{D_L}}(\zeta|\mathbf{D_L}) - \tau(\zeta) \mid > \epsilon) = 0
\]
where $\tau(\zeta) = (2\pi)^{-\frac{p}{2}} \exp(-\frac{1}{2} \|\zeta\|^2)$.
\end{thm}

Theorem~\ref{BVM} asserts that for a smooth finite-dimensional model with an assumption that the prior is continuous and has positive mass in the neighborhood around true parameter $\theta_*$, as the number of observations tends to infinity, the posterior distribution of a parameter is approximately a normal distribution~\cite{schervish2012theory,van2000asymptotic, 10.1214/12-EJS675}. 
In other words, as $n\rightarrow \infty$, Bayesian credible intervals are asymptotically equivalent to the Wald confidence sets~\cite{van2000asymptotic} based on the asymptotically normal estimator (i.e., the MLE) which makes them valid confidence sets. As a result the posterior inferences are asymptotically correct in the frequentist sense. 

Theorem~\ref{BVM} ensures that the derived posterior $P^*$ 
converges to a normal distribution with the mean of true parameter, as the number of observations increases. 
In light of this convergence, we will show the \textbf{Bias} term vanishes so that the \textbf{Variance} term will faithfully capture MSE and represent the model's prediction performance. More specifically, considering that for a finite-dimensional estimator in the same parameter space of the true parameter, as the number of observations increases, $\pi^*$ converges to $P(Y|\mathbf{X})$, we show \textbf{Variance} can be an effective acquisition function leading to a better UAL performance (lower MSE and thereafter $\text{MSE}_\text{obs}$).

Hence, the following proposition is proved for the vanishing bias when we have the prediction model class covering the underlying ground truth. 

\begin{proposition}\label{prop:bias}
    $(E_{P^*(\theta|\mathbf{D_L})}[\hat{f}_{\theta}]-  E_{P(Y|\mathbf{X})}[Y])^2 < \varepsilon^2 C^2$ if $E_{P(Y|\mathbf{X})}[|Y|]<C$ and $\big|\frac{\pi^*}{P(Y|\mathbf{X})}-1\big|<\varepsilon$, where $C > 0$ and $\varepsilon > 0$ are constants.
    \footnote{Proof is provided in the Appendix.}
\end{proposition}

Proposition \ref{prop:bias} indicates that as $\varepsilon$ goes to $0$, \textbf{Bias} goes to $0$ as well.

Although Theorem~\ref{BVM} only considers parameterized models, in section \ref{experiments}, we empirically show the \textbf{Variance} is indeed an effective acquisition function and represents the MSE when the underlying predictive model covers the true distribution for both parametric and non-parametric models. 

Having the MSE decomposed into the bias and variance terms, the next step is to provide illustrative UAL performance analyses for the corresponding predictive model, $\hat{f}_{\theta}$. 
We choose to 
concentrate on Bayesian Regression with polynomial regressors as predictive models. 
This choice of prediction model classes leads to closed-form MSE analysis and also enables straightforward model class capacity manipulation by increasing the polynomial order.


Bayesian Polynomial Regression~(BPR) is a widely utilized 
parametric model class with 
polynomial functions. 
Let $\hat{f}_{\theta}(\mathbf{x}) = \langle{\phi(\mathbf{x},p)}, \theta\rangle$, where $\phi(\mathbf{x},p)$ 
is a non-linear operator that maps $\mathbf{x}$ to a $p^\text{th}$-order polynomial expansion 
and $\theta$ 
is the corresponding weight vector~\cite{murphy2012machine}. 

With the typical white noise residual error assumption, the observed output $Y$ follows the Gaussian distribution, 
\begin{equation}
    P(Y|\mathbf{X}=\mathbf{x}) = N(Y|f(\mathbf{x}), \sigma^2).
\end{equation}
We can derive the predictive posterior given $\mathbf{x}$ 
in BPR 
with the conjugate prior.  
Specifically, with the conjugate prior $\theta \sim N(\hat{\mu}, \hat{\Sigma})$ where $\hat{\mu}$ 
and $\hat{\Sigma}$ 
are the prior mean and covariance, the posterior of the model's parameters can be derived by Bayes' rule: 
\begin{equation}\label{eq:4}
\begin{aligned}
    P^*(\theta|\hat{\mathbf{\Phi}},{\mathbf{y}},\sigma^2) & \propto N(\theta|\hat{\mu}, \hat{\Sigma})N({\mathbf{y}}|\hat{\mathbf{\Phi}}\theta, \sigma^2)\\
    &\propto N(\theta|\hat{\mu}_p, \hat{\Sigma}_p), \\
    \hat{\mu}_p &= \hat{\Sigma}_p\hat{\Sigma}^{-1}\hat{\mu} + \frac{1}{\sigma^2}\hat{\Sigma}_p\hat{\mathbf{\Phi}}^\top {\mathbf{y}}; \\
    \hat{\Sigma}_p^{-1} &= \hat{\Sigma}^{-1} + \frac{1}{\sigma^2}\hat{\mathbf{\Phi}}^\top \hat{\mathbf{\Phi}}, 
\end{aligned}
\end{equation}
where $\hat{\mu}_p, \hat{\Sigma}_p$ denote the posterior mean and covariance respectively, $\hat{\mathbf{\Phi}}=[\phi_1^p, \dots,\phi_n^p]^\top $ is the transformed input matrix with $n$ transformed $p^{th}$-order polynomial terms 
$\phi_i^p = \phi(\mathbf{x}_i,p)$, 
and ${\mathbf{y}} = [{y}_1, \dots, {y}_n] \in \mathbb{R}^{n}$ is the corresponding training output vector with ${y}_i \sim P(Y|\mathbf{X}=\mathbf{x}_i)$.   

By substituting the parameter posterior to Eq. \eqref{eq:predctive_posterior}, It can be shown that the predictive posterior at $\mathbf{x}$ follows a Gaussian distribution: 
\begin{equation}\label{eq:5}
    \pi^*(\hat{y}|\mathbf{x}) = N(\hat{y}|\langle\phi(\mathbf{x},p),\hat{\mu}_p\rangle, \sigma_p^{2}(\mathbf{x})), 
\end{equation}
where
\begin{equation}\label{eq:6}
    \sigma^{2}_p(\mathbf{x}) = \sigma^2 + \phi(\mathbf{x}, p)^\top \hat{\Sigma}_p\phi(\mathbf{x}, p)
\end{equation}
is the posterior predictive variance. Since $\sigma_p^2(\mathbf{x})$ captures model prediction uncertainty and following the Proposition~\ref{prop:bias}, 
it can be used to define a meaningful acquisition function to guide UAL for BPR. 
As a result, in this study, we consider posterior predictive variance in~\eqref{eq:6} as the acquisition function, i.e., $\mathcal{A}(\pi^*, \mathbf{x}) = \sigma_p^2(\mathbf{x})$, and the following analysis is based on this choice.

Without loss of generality, to evaluate the reduction of the MSE of actively learned models given acquired labeled training data, we consider 
BPR with univariate inputs ($d=1$) with $\phi(\mathbf{x},p) = [1, \dots, x^{p}]$ ($p$: the polynomial order, considered as the model complexity). 
We note that the analysis results can be easily generalized to multivariate cases.

As mentioned earlier in this section, when the derived posterior 
is close enough to the ground truth model that generates the data, \textbf{Variance} can capture the MSE, 
thus leading UAL to a better performance. 
This closeness is affected by the prediction model class capacity. 
More specifically, when the model class capacity is flexible enough to capture the underlying ground truth function (i.e., when the model class is in the same parameter space as the ground truth or it covers the ground truth's parameter space), the MSE can be estimated based on the prediction \textbf{Variance}.

\begin{figure}[htbp]\label{fig: gp_mot}

   \centerline{\includegraphics[width=1.02\linewidth]{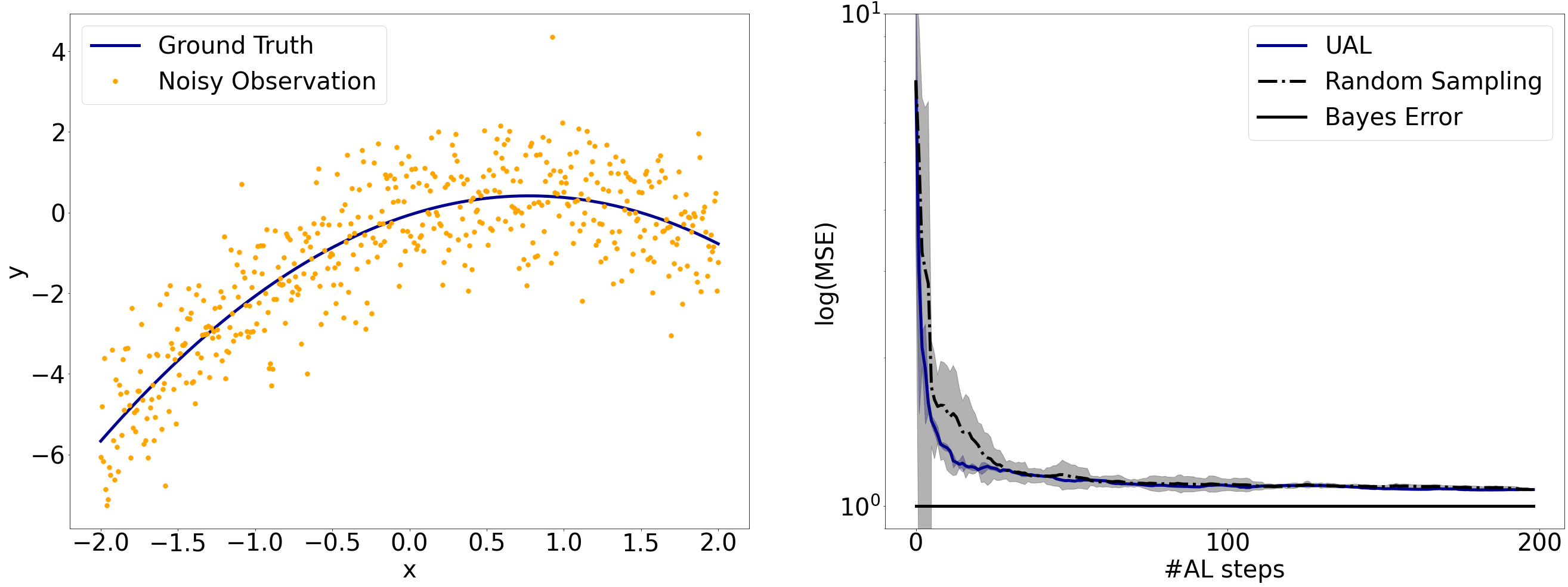}}
    \caption{Motivating example with a ground truth target that is as complex as the prediction model class. In this example, the noisy data follows the ground truth function which is a quadratic function ($y = f(x)+\epsilon$ with $\epsilon \sim N(0,1)$, where $f(x)=\langle\mathbf{x},\mathbf{w}\rangle$, and $\mathbf{x} = [1,x,x^2]$). The left plot shows the ground truth target function with the same $\mathbf{w}$ as the example in Section \ref{Introduction} and its corresponding noisy observed data. The right plot shows the comparison of UAL and random sampling performance to learn the quadratic predictor based on Bayesian polynomial regression. \\ 
    }
    \label{fig: Fig(motMT)}
\end{figure}

As an example, Figure \ref{fig: Fig(motMT)} compares the UAL performance with random sampling when the underlying function's complexity matches the predictive model class with both being quadratic functions. It is clear that UAL is able to outperform random sampling, implying that, aligned with our analysis, the estimated acquisition function based on the predictive variance faithfully 
represents 
MSE when the model class covers the ground truth target, which leads to efficient UAL.

\begin{figure}[htbp]\label{fig: gp_mot}

   \centerline{\includegraphics[width=0.7\linewidth]{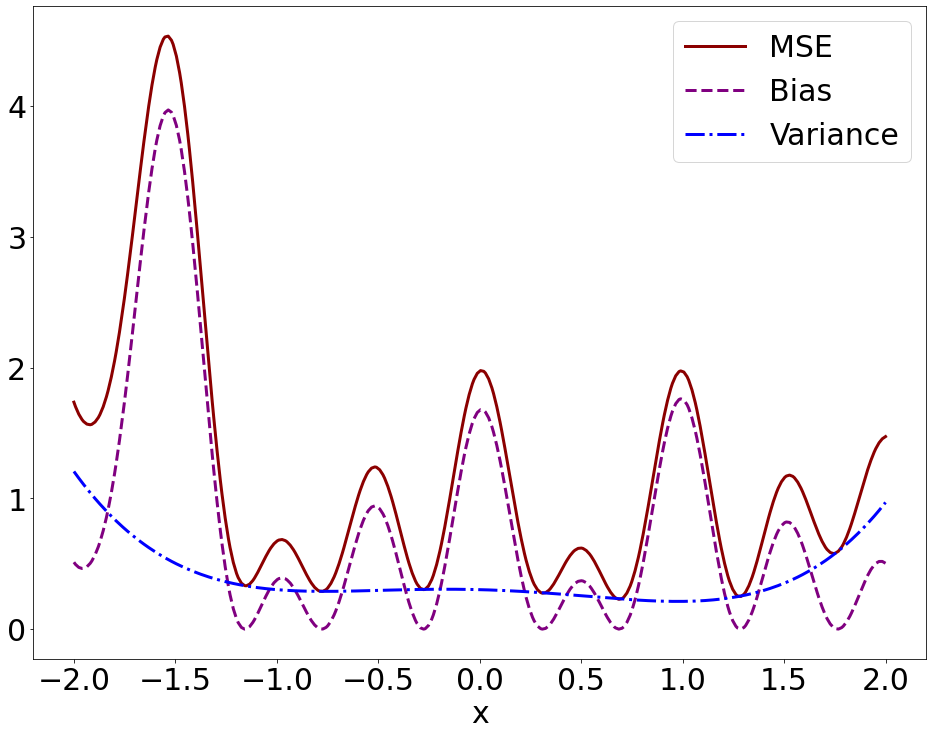}}
    \caption{Bias-variance decomposition for the motivating example with lower complexity prediction model in Section \ref{Introduction}. In this example with the incapable prediction model compared to the target function, the estimated variance cannot capture the learning objective, MSE.}
    \label{fig: gp_mot_bvd}
\end{figure}

While when checking the bias and variance decomposition of the experiment in Section \ref{Introduction} as demonstrated in Figure \ref{fig: gp_mot_bvd}, since the model class complexity is not enough to cover the target function, the bias term can dominate the MSE and  the variance term does not fully capture the MSE, making it an uninformative acquisition function unable to guide UAL effectively.

In what follows, we delve deeper into the analysis of MSE, considering an uncertainty class of the ground truth functions. 
Assuming that the uncertainty class of ground truth functions is the $l^{th}$-order polynomial family, 
$f(\mathbf{x}) = \langle\phi(\mathbf{x},l), \mathbf{w}\rangle \in \mathbb{R}$, with 
$\mathbf{w}\sim N(\mu, \Sigma)$, the MSE can be decomposed as: 
\[
\begin{aligned}
\text{MSE}&= E_{P(Y|\mathbf{X})}\big[E_{P^*(\theta|\mathbf{D_L})}[(f(\mathbf{x})-\langle\phi(\mathbf{x},p),\hat{\mu}_p\rangle)^2] \\
        &\quad+ E_{P^*(\theta|\mathbf{D_L})}[(\langle\phi(\mathbf{x},p),\hat{\mu}_p\rangle-\langle\phi(\mathbf{x},p),\theta\rangle)^2]\big] \\
    &= E_{P(Y|\mathbf{X})}[(f(\mathbf{x})-\langle\phi(\mathbf{x},p),\hat{\mu}_p\rangle)^2 \\
    &\quad+ \phi(\mathbf{x},p)^\top \hat{\Sigma}_p\phi(\mathbf{x},p)]\\
    &= \underbrace{E_{P(\mathbf{w})}[E_{P(Y|\mathbf{w},\mathbf{X})}[(f(\mathbf{x})-\langle\phi(\mathbf{x},p),\hat{\mu}_p\rangle)^2]]}_{\mbox{\textbf{Bias}}} \\
    &\quad+ \underbrace{\phi(\mathbf{x},p)^\top \hat{\Sigma}_p\phi(\mathbf{x},p)}_{\mbox{\textbf{Variance}}}. 
\end{aligned}
\]

By replacing posterior mean ($\hat{\mu}_p$) in~\eqref{eq:4} and taking the expectation, the MSE is: 
\begin{equation}{\label{eq:12}}
    \begin{aligned}
    \text{MSE} &=  (\phi^l)^\top (\mu\mu^\top  + \Sigma)(\phi^l) - 2(\phi^p)^\top \hat{\Sigma}_p\hat{\Sigma}^{-1}\hat{\mu}\mu^\top \phi^l \\
    &\quad- \frac{2}{\sigma^2}(\phi^p)^\top \hat{\Sigma}_p\hat{\mathbf{\Phi}}^\top \mathbf{\Phi}(\mu\mu^\top +\Sigma)\phi^l \\
    &\quad +(\phi^p)^\top \hat{\Sigma}_p\hat{\Sigma}^{-1}\hat{\mu}\hat{\mu}^\top  \hat{\Sigma}^{-1}\hat{\Sigma}_p(\phi^p) \\
    &\quad+ \frac{1}{\sigma^2} (\phi^p)^\top \hat{\Sigma}_p\hat{\Sigma}^{-1}\hat{\mu}\mu^\top \mathbf{\Phi}^\top \hat{\mathbf{\Phi}}\hat{\Sigma}_p(\phi^p) \\
    &\quad+ \frac{1}{\sigma^2}(\phi^p)^\top \hat{\Sigma}_p\hat{\mathbf{\Phi}}^\top \mathbf{\Phi}\mu\hat{\mu}^\top \hat{\Sigma}^{-1}\hat{\Sigma}_p(\phi^p) \\
    &\quad+ \frac{1}{\sigma^4}(\phi^p)^\top \hat{\Sigma}_p\hat{\mathbf{\Phi}}^\top \mathbf{\Phi}(\mu\mu^\top +\Sigma)\mathbf{\Phi}^\top \hat{\mathbf{\Phi}}\hat{\Sigma}_p(\phi^p)\\
    &\quad+\frac{1}{\sigma^2}(\phi^p)^\top \hat{\Sigma}_p\hat{\mathbf{\Phi}}^\top \hat{\mathbf{\Phi}}\hat{\Sigma}_p(\phi^p) + (\phi^p)^\top \hat{\Sigma}_p(\phi^p),
    \end{aligned}
\end{equation}
where $\phi^p =\phi(\mathbf{x},p)\text{, } \phi^l = \phi(\mathbf{x},l)$, and $\mathbf{\Phi} \in \mathbb{R}^{n\times (l+1)}$.

Assuming the model's complexity matches the target's complexity ($p=l$), we can further simplify the MSE. More specifically, based on this assumption, $\mathbf{\Phi}$ and $\hat{\mathbf{\Phi}}$, $\phi^l$  and $\phi^p$ are equivalent. Furthermore, knowing the uncertainty class of ground truth functions, $\mu$ and $\hat{\mu}$, $\Sigma$ and $\hat{\Sigma}$ are equal. As a result, $\mathbf{\Phi}\hat{\mathbf{\Phi}}^\top =\hat{\mathbf{\Phi}}\mathbf{\Phi}^\top =\hat{\mathbf{\Phi}}\hat{\mathbf{\Phi}}^\top $. Also, considering that $\hat{\mathbf{\Phi}}\hat{\mathbf{\Phi}}^\top  = \sigma^2 (\hat{\Sigma}_p^{-1} - \hat{\Sigma}^{-1})$, by replacing $\hat{\mathbf{\Phi}}\hat{\mathbf{\Phi}}^\top $ with its equivalent, the MSE for the matched model is\footnote{The detailed derivation for Eq. \eqref{eq:12} and \eqref{eq:14} is provided in Appendix.}:

\begin{equation}{\label{eq:13}}
    \begin{aligned}
    \text{MSE} &= 2(\phi^p)^\top \hat{\Sigma}_p(\phi^p) = 2 (\sigma_p(\mathbf{x})^2-\sigma^2)    
    \end{aligned}
\end{equation}

Eq. \eqref{eq:13} shows that when the model matches, the \textbf{Variance} 
is proportional to the MSE, i.e., using the posterior predictive variance, one can find what would be the MSE of a sample for the currently trained model, and consequently, which point is the best to be added to the training set. This can potentially help UAL outperform the random sampling.

With Eq. \eqref{eq:13} giving an intuitive explanation to MSE of models with matched complexity as the target, the analysis for the unmatched models would not lead to a similar clean form. 
Resting upon Theorem~\ref{BVM}, with reasonable Bayesian inference, higher-order prediction models in BPR can provide a predictive variance that still captures the actual MSE fairly well. 
However, the lower-order model exhibits a notably pronounced \textbf{Bias} with respect to its \textbf{Variance}, 
deterring the effectiveness of using the \textbf{Variance} as the MSE surrogate. For this reason, we further analyze the lower-order model where the model's order is smaller than the target's order ($p<l$).

Before delving into the analysis, we define the following matrices: 
\[
\begin{aligned}
    \mathbf{\Phi} =& [\tilde{\mathbf{\Phi}}_c \quad \hat{\mathbf{\Phi}}] \in \mathbb{R}^{n\times (l+1)}, \qquad
    \mu = \begin{bmatrix}
    \tilde{\mu}_{c} \\  \tilde{\mu}
    \end{bmatrix} \in \mathbb{R}^{(l+1)} \\
    \phi^l =& \begin{bmatrix}
        \tilde{\phi}_c \\ \phi^p 
    \end{bmatrix} \in \mathbb{R}^{(l+1)} ,\qquad \quad
    \Sigma = \begin{bmatrix}
        \tilde{\Sigma}_{c} \quad \Sigma_{12}\\ \\
        \Sigma_{12}^\top  \quad \tilde{\Sigma}
    \end{bmatrix} \in \mathbb{R}^{(l+1)\times (l+1)} 
\end{aligned}
\]
with $\tilde{\phi}_c = [x^{p+1}, \dots, x^{l}]$, where 
\[
\begin{aligned}
     \tilde{\mathbf{\Phi}}_c \in \mathbb{R}^{n\times(l-p)}, &\qquad \hat{\mathbf{\Phi}} \in \mathbb{R}^{n\times (p+1)}, \\
     \tilde{\mu}_{c} \in \mathbb{R}^{(l-p)}, &\qquad \tilde{\mu} \in \mathbb{R}^{ (p+1)} \\
     \tilde{\phi}_c \in \mathbb{R}^{(l-p)}, &\qquad \phi^{p} \in \mathbb{R}^{(p+1)}\\
    \tilde{\Sigma}_c \in \mathbb{R}^{(l-p)\times (l-p)},  \Sigma_{12} &\in \mathbb{R}^{(l-p)\times (p+1)},  \tilde{\Sigma} \in \mathbb{R}^{(p+1)\times (p+1)}
    \\
\end{aligned}
\]

Assuming $\tilde{\mu} = \hat{\mu}$ and $\tilde{\Sigma}=\hat{\Sigma}$, after plugging in the equivalent matrices to Eq. \eqref{eq:12}, the MSE will be:
\begin{equation}\label{eq:14}
\begin{aligned}
\text{MSE} &= (\tilde{\phi}_c)^\top (\tilde{\Sigma}_{c}+\tilde{\mu}_{c}\tilde{\mu}_{c}^\top )(\tilde{\phi}_c) \\
& \quad-\frac{2}{\sigma^2} (\phi^p)^\top \hat{\Sigma}_p\hat{\mathbf{\Phi}}^\top \tilde{\mathbf{\Phi}}_c(\tilde{\Sigma}_{c}+\tilde{\mu}_{c}\tilde{\mu}_{c}^\top )(\tilde{\phi}_c)\\
& \quad+2(\phi^p)^\top \hat{\Sigma}_p\hat{\Sigma}^{-1}\Sigma_{12}^\top (\tilde{\phi}_c)\\
&\quad +\frac{1}{\sigma^4} (\phi^p)^\top \hat{\Sigma}_p\hat{\mathbf{\Phi}}^\top \tilde{\mathbf{\Phi}}_c(\tilde{\Sigma}_{c} + \tilde{\mu}_{c}\tilde{\mu}_{c}^\top )\tilde{\mathbf{\Phi}}_c^\top \hat{\mathbf{\Phi}}\hat{\Sigma}_p(\phi^p) \\
&\quad- \frac{2}{\sigma^2} (\phi^p)^\top \hat{\Sigma}_p \hat{\mathbf{\Phi}}^\top \tilde{\mathbf{\Phi}}_c\Sigma_{12}\hat{\Sigma}^{-1}\hat{\Sigma}_p(\phi^p) \\
&\quad + 2(\phi^p)^\top \hat{\Sigma}_p(\phi^p) = \mathrm{P}(\mathbf{x}) + 2\times \text{Var}(\mathbf{x}), 
\end{aligned}
\end{equation}
where the \textbf{Variance} term $\text{Var}(\mathbf{x})= (\phi^p)^\top \hat{\Sigma}_p(\phi^p)$ is a $(2\times p)^{th}$-order polynomial function of $\mathbf{x}$, and $\mathrm{P}(\mathbf{x})$ consisting of the remaining terms in $\text{MSE}$ is a $(2\times l)^{th}$-order polynomial function of $\mathbf{x}$.

Importantly, it means in the lower-order model case, the complexity (order) of the \textbf{Variance} term diverges from the true MSE. This would potentially result in the inability of the estimated uncertainty to accurately capture the true learning objective, MSE, as we will observe in the later experiments. 
Consequently, relying on \textbf{Variance} (uncertainty metric) for sample selection yields choices that lack informativeness.

\section{Effective UAL for Regression }\label{remedy}
Hitherto, we have explained through theoretical analysis and simulations that when the predictive model family is incapable of covering the underlying ground truth, acquired uncertainties 
may lead to impractical acquisition functions and, hence, an unhelpful UAL procedure. 
In light of the results of Section~\ref{UAL efficacy}, here we aim to investigate potential avenues to address the drawbacks of UAL under model mismatch scenarios. In particular, we approach this problem by attending to what to consider in the design of the acquisition function.


Our derived insight that variance does not reflect learning objective(s) under model mismatch lays the foundation for what should be considered in the design of an effective acquisition function. 
The key is to represent the true objective function for any unseen sample based on the available labeled dataset to guide UAL. 
%
We focus on the regression task, where MSE is the learning objective, and thus, the design of the acquisition function should be dependent on that. In this part, we describe two ways to design such acquisition functions. 

One path is similar as Krigging~\cite{cressie1990origins} to leverage another model that directly estimates the MSE. In this strategy, at each step of the UAL process, a secondary estimator is trained simultaneously on the labeled dataset, $\mathbf{D_L}$, which then is utilized to estimate the MSE and thereafter, to design the acquisition function. Let $g: \mathcal{X} \rightarrow \mathbb{R}$ as the secondary model estimating the ground truth, $f$, the label of selected $\mathbf{x}^* \in \argmax_{\mathbf{x_U}\in \mathbf{D_U}} \mathcal{A}(\theta, \mathbf{x_U})$ is queried and appended to the labeled dataset, where $\mathcal{A}(\theta,\mathbf{x}) = (g(\mathbf{x})-f_\theta(\mathbf{x}))^2$ is the acquisition function.


Noting that to select the new sample from the unlabeled dataset at each step in UAL, only the relative trend of the objective function matters, 
the other path that we can take is designing an acquisition function that faithfully captures the objective's trend to help prioritize, which can leverage recent advancements in error estimators or error bounds. 
In essence, let $\tilde{g}:\mathcal{X} \rightarrow \mathbb{R}$ as the estimated error upper bound where $\tilde{g}(\mathbf{x}) \geq e(\mathbf{x}, \theta), \text{ for all } \mathbf{x}\in \mathcal{X}$, with $e(\mathbf{x}, \theta) = (y-\hat{f}_{\theta}(\mathbf{x}))^2$, the new unlabeled sample $\mathbf{x}^*$ is selected using $\mathcal{A}(\theta, \mathbf{x}) = \tilde{g}(\mathbf{x})$ in the same manner as the previous solution. 

In Section~\ref{experiments}, we provide an example for each approach and empirically demonstrate their efficacy.

\section{Experiments}\label{experiments}

We have theoretically analyzed the performance of UAL algorithms, which  relies on whether their adopted acquisition function based on the estimated uncertainty 
can faithfully portray the ultimate learning objective, MSE, when considering regression in this paper.  
In Section~\ref{UAL efficacy}, we show that 
the predictive variance 
is able 
to capture the regression learning objective when the model's capacity aligns or surpasses that of the underlying ground truth target function.

In this section, to further investigate the validity of our findings and the effect of model complexity mismatch on UAL performance, we provide a more comprehensive evaluation of experimental results {by utilizing a synthetic dataset tailored to our aims} to demonstrate the relationships between the predictive model complexity, acquisition function effectiveness, and UAL performance (measured by MSE). 
{To further justify the reliability of our results, we carry out experiments to demonstrate variance-based UAL performance on two real-world open-access datasets comparing the UAL performances with a simple model and a complex model representing low vs. high model complexity classes respectively.}
{Finally}, to overcome the shortcomings of UAL with incapable predictive models, building upon the insights provided in Section~\ref{remedy}, we recommend using MSE-dependent acquisition functions for UAL. Specifically, we showcase the effectiveness of two simple MSE-dependent solutions by providing experiments using the same target used in Section~\ref{Introduction}.

\subsection{UAL by Estimated Variance}\label{IV:A}

The UAL method in all our experiments and simulations 
is based on the basic pool-based UAL setting: the prediction model will be trained on the initial labeled dataset, $\mathbf{D_L}$; the most uncertain sample based on the learned model from the unlabeled dataset $\mathbf{D_U}$ will then be queried and added to the labeled dataset to update the model in the next step. This process iterates and ends when all the designated unlabeled training data, $\mathbf{D_U}$, is exploited. In our experiments for UAL with regression, the predictive posterior variance is considered as the measure of uncertainty, and considered as the acquisition function to select new samples from $\mathbf{D_U}$. 

For deriving generalizable conclusions, we conduct experiments 
{on a synthetic dataset that consists of} 100 {randomly generated} $3^{rd}$-order polynomial functions as the ground truth functions to simulate the observed data for UAL performance evaluation. We collect the corresponding performance statistics from these 100 random runs to illustrate the influence of model complexity mismatch on UAL performance. For each run, we first generate an unlabeled training dataset ($\mathbf{D_U}$) of 200 evenly partitioned samples with $x \in [-2, 2]$. We then simulate the output corresponding to each selected sample 
$x_n$ during the UAL procedure based on the following noisy observations: 
\begin{equation}\label{noisy_groundTruth}
    \begin{split}
        {{y}_n} &= {f}_n + \epsilon, \quad\quad
        \epsilon \sim N(0,\sigma^2), 
    \end{split}
\end{equation}
where $f_n = \langle(\phi_n^3)^\top , \mathbf{w}\rangle$ with $(\phi_n^3)^\top  = [1, x_n, x_n^2, x_n^3]$ and the corresponding randomly sampled coefficient vector for the $3^{rd}$-order polynomial $\mathbf{w} \in \mathbb{R}^{4}$, $\mathbf{w}\sim N(\textbf{0}, \textbf{I})$, and $\sigma=1$. 
For each run, the initial labeled dataset ($\mathbf{D_L}$) consists of one sample $x$ that is randomly drawn from $\mathbf{D_U}$ and its corresponding label $y$ that is acquired from Eq. \eqref{noisy_groundTruth} with the specific coefficient vector $\mathbf{w}$. 
To evaluate the regression and UAL performance, a holdout test dataset of 500 samples with $x\in[-2,2]$ and their corresponding 
observed output $y$ is used to estimate the test MSE. 

In the following, we perform two sets of experiments {on this synthetic dataset} to check the performance of UAL when using Bayesian Polynomial and Gaussian Process Regressors as the predictive model, respectively.  The initial $\mathbf{D_L}$ and target functions are the same across both sets.

{To verify the reliability of our conclusions, we provide additional experimental results on two real-world case studies. In particular, we use the \emph{Concrete Compressive Strength} \cite{misc_concrete_compressive_strength_165} and \emph{Facebook Metrics} \cite{misc_facebook_metrics_368} datasets for the regression task. For the \emph{Facebook Metrics} dataset, we use 25\% of the dataset as the test dataset and the rest as the training dataset. For the \emph{Concrete Compressive Strength}, we randomly select 600 samples from the dataset, with the same train/test split ratio as in the \emph{Facebook Metrics} dataset. In both case studies, we consider using Gaussian Process Regressors as the predictive model.}

\subsubsection{Bayesian Polynomial Regression}\label{IV:B}

In this set of experiments, we {perform Bayesian polynomial Regression~(BPR) for the corresponding model class of polynomial order one (linear) to five and compare the learning performance based on the UAL and random sampling strategies.} In each run, starting from one randomly selected pair of input and output, the performance of each polynomial model in UAL and random sampling procedure is evaluated.

\begin{figure*}
    \centering
    \includegraphics[width = 1\linewidth]{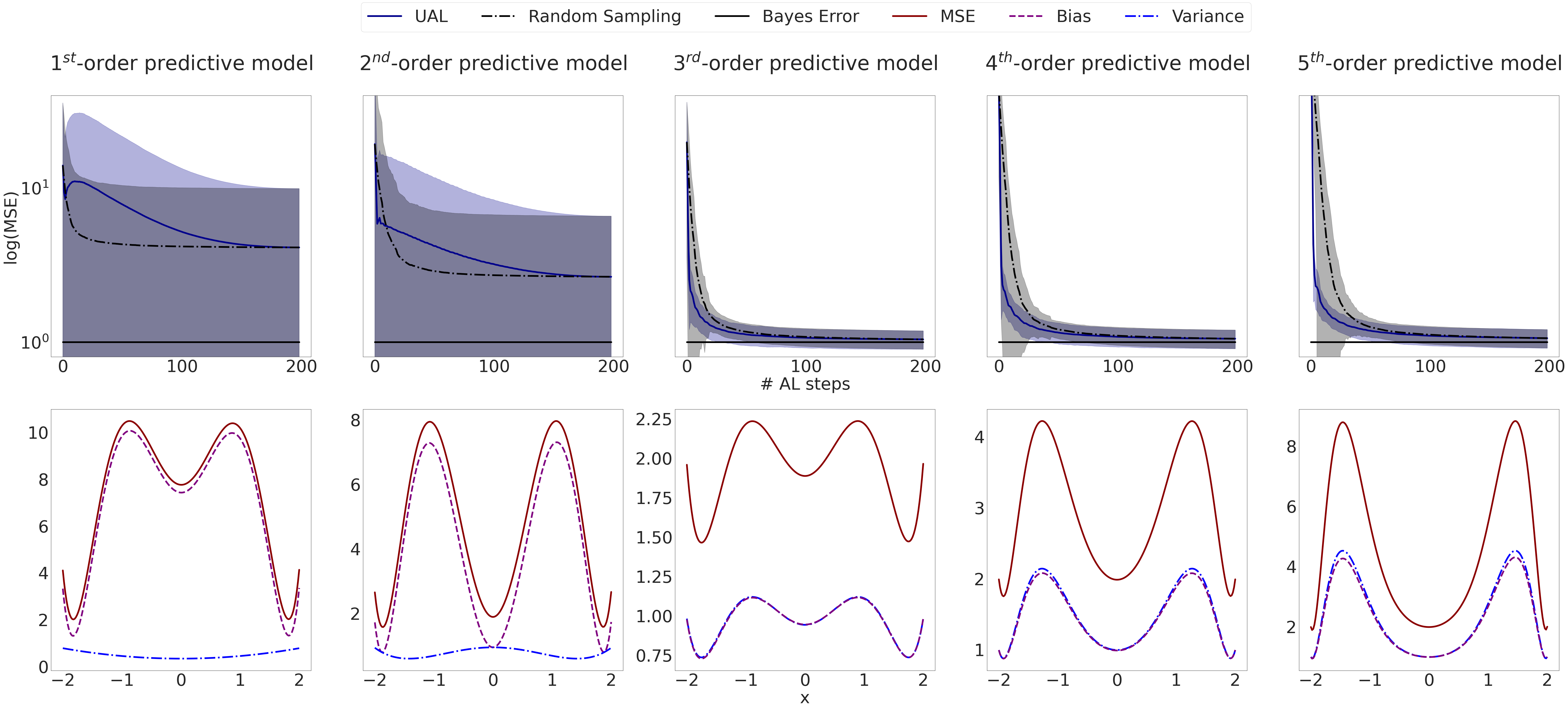}
    \caption{Performance of model class of polynomial order one to five and their corresponding bias-variance decomposition on BP regression task over the $3^{rd}$-order polynomial family. 
    The first row shows the performance and the second row shows the bias-variance decomposition related to each model. As the prediction model becomes more complicated, the ability of variance to capture the learning objective (MSE) increases; hence, UAL performance improves.}
    \label{fig:BPR_alrnd_bvd}
\end{figure*}

The striking trend in Figure~\ref{fig:BPR_alrnd_bvd} is that when the model class is linear or quadratic functions, UAL shows significant worse performance than random sampling, while when the model class complexity increases over the ground truth $3^{rd}$-order polynomial functions, UAL performs better as typically expected. 
This can be explained based on the analysis in Section \ref{UAL efficacy}.

To clearly illustrate the issues of UAL when the model class does not cover the target ground truth function to learn, we further visualize the \textbf{Bias} and \textbf{Variance} decomposition 
patterns in the second row of Figure \ref{fig:BPR_alrnd_bvd}. It is clear that when the model class complexity matches or surpasses the target's complexity, the estimated predictive variance captures the actual MSE. However, model classes (linear and quadratic) with low capacity are incapable of providing an accurate estimation of MSE via the variance-based UAL acquisition function; hence, they lead to degraded UAL performance, which can perform even worse than random sampling. 

\begin{figure}
    \centering
    \includegraphics[width = 1\linewidth]{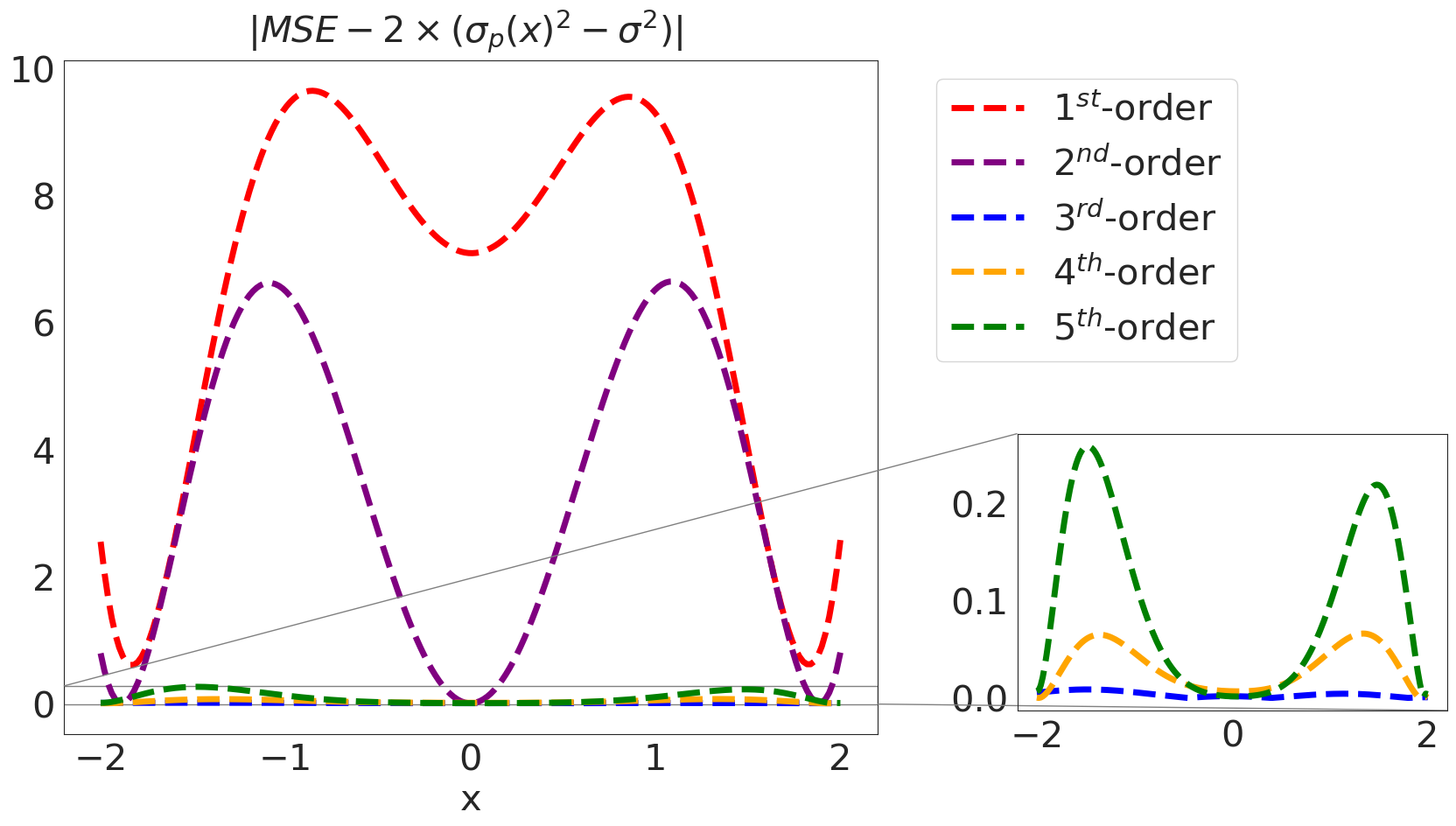}
    \caption{The absolute difference between $\text{MSE}$ and $2\times (\sigma_p^2(\mathbf{x})-\sigma^2)$. Aligned with the analysis in Section~\ref{UAL efficacy}, in BPR, when the prediction model's complexity matches the target function's complexity, the MSE becomes $2\times (\sigma_p^2(\mathbf{x})-\sigma^2)$ (Eq.~\eqref{eq:13}). However, for mismatched models, both with higher and lower order models, this equation does not hold, with a striking difference for lower order models. }
    \label{fig:bvddiff}
\end{figure}

Figure~\ref{fig:bvddiff} further illustrates the effectiveness of the estimated predictive variance in representing the MSE by showing the absolute difference between $\text{MSE}$ and $2\times (\sigma_p^2(\mathbf{x})-\sigma^2)$ (Eq.~\eqref{eq:13}). 
There is a clear difference between actual $\text{MSE}$ and $2\times (\sigma_p^2(\mathbf{x})-\sigma^2)$ for mismatched models as expected. 
However, when the complexity of the model class surpasses targets,  we observe \textbf{Bias} gradually decreases and \textbf{Variance} captures the true MSE, which makes it capable of guiding UAL. 
The decreasing trending of the \textbf{Bias} may be related to the recent `double descent' observations for Neural Networks that when the model is highly over-parameterized, it may considered to generalize well.

Another observation in the 
BPR experiment is that in the early stages of the UAL a much sharper drop in the error happens compared to random sampling.
Even for models with insufficient complexity, this phenomenon happens before they get stuck in the uninformative regions. 
However, after passing the early stages of UAL and observing more samples, 
random sampling will outperform UAL.

In general, all UAL models performed better than random sampling at the very beginning of BPR, but as the UAL process progressed, the models whose complexity fall short of the target's complexity got stuck in the uninformative regions of the input space leading to significantly poor performance of these UAL models compared to random sampling.

\begin{figure}[htbp]\vspace{-5mm}
    \centering
    \includegraphics[width = 0.65\linewidth]{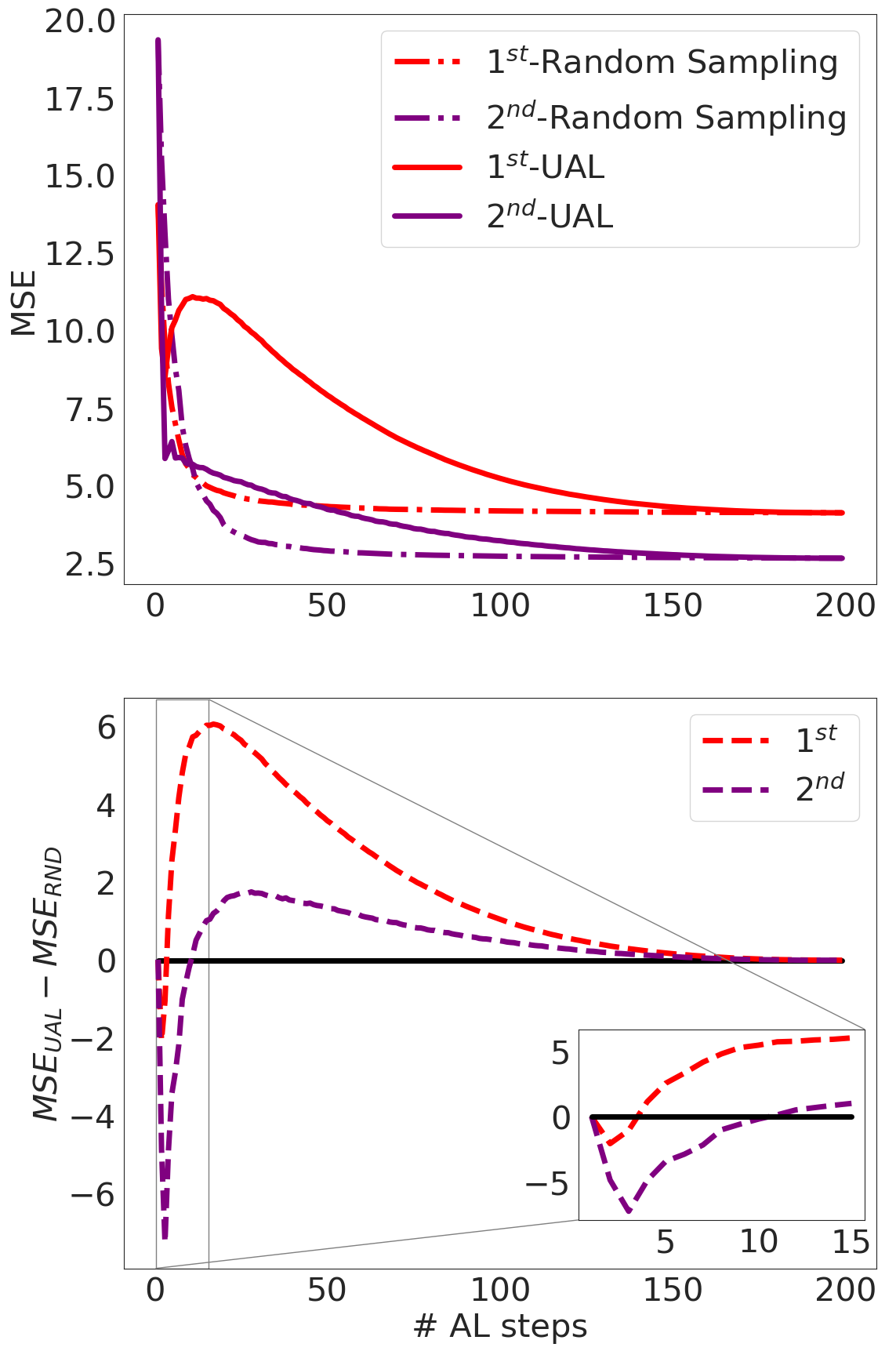}
    \caption{UAL error versus random sampling error for the model class with lower capacity in BPR experiments. The {top} figure demonstrates when the prediction models' complexity is lower than the ground truth target function's complexity. After the early steps of UAL, it gets stuck in querying non-informative regions of the input space. The {bottom} figure illustrates the difference between UAL and random sampling MSE. As shown, in BPR experiments, even models with insufficient complexity have superior performance juxtaposed to random sampling at the early steps of UAL.}
    \label{fig:4}
\end{figure}

\subsubsection{Gaussian Process Regression}\label{IV:C}

{In this} set of experiments, we {first} perform Gaussian Process~(GP) regression {on the synthetic dataset} to further validate the generalizability of our findings with more flexible prediction models. {Then we conduct the regression task on two real-world datasets to verify the reliability of our conclusions.}

Instead of modeling the correlation of $\mathbf{X}$ and $Y$ with parameters, GP assumes the outputs are jointly Gaussian distributed with a mean function $m(\mathbf{x})$ and covariance defined by a kernel function $k(\mathbf{x}, \mathbf{x}')$, i.e. $\emph{\textbf{f}}(\mathbf{x}) \sim \mathcal{GP}(m(\mathbf{x}), k(\mathbf{x}, \mathbf{x}'))$. The conditional distribution $P(\emph{\textbf{f}}(\mathbf{x})|\mathbf{D_L}) = \mathcal{N}(m_\mathbf{D_L}(\mathbf{x}), \sigma^2_\mathbf{D_L}(\mathbf{x}))$ is considered to be prediction, where the mean $m_\mathbf{D_L}(\mathbf{x}) = m(\mathbf{x}) + \textbf{k}_*^\top  (\textbf{K} + \sigma^2 \textbf{I})^{-1} (\mathbf{y} - \textbf{m})$, the variance $\sigma^2_\mathbf{D_L}(\mathbf{x}) = k(\mathbf{x}, \mathbf{x}) - \textbf{k}_*^\top  (\textbf{K} + \sigma^2 \textbf{I})^{-1} \textbf{k}_*$, $\textbf{k}_*  = [k(\mathbf{x}, \mathbf{x}_1), \dots, k(\mathbf{x}, \mathbf{x}_n)]^\top $, $\textbf{m} = [m(\mathbf{x}_1), \dots, m(\mathbf{x}_n)]^\top $, the covariance matrix of the observed points $\textbf{K}  = [\textbf{k}_1, \cdots \textbf{k}_n]$, and $\textbf{k}_i  = [k(\mathbf{x}_i, \mathbf{x}_1), \dots, k(\mathbf{x}_i, \mathbf{x}_n)]^\top $.

The selection of the kernel function $k(\mathbf{x}, \mathbf{x}')$ plays a critical role in Gaussian Process Regression~(GPR) tasks.
When choosing linear kernels, GPR is equivalent to BPR with the linear model class. 
As the kernel becomes more complicated, the model's capacity increases accordingly. 
{Similar to the previous set, for experiments on both synthetic and real-world datasets, GPR with a linear kernel representing low-capacity models is tested. During the experiments on the synthetic dataset, we choose the Matern kernel to account for complex models. For the experiments on real-world datasets, we exploit GPR with a Radial Basis Function (RBF) kernel to represent high-capacity models.}

Using the same {synthetic} 100 $3^{rd}$-order ground truth polynomial functions, 
as in the BPR experiments, each run starts with one randomly selected pair of input and output corresponding to one $3^{rd}$-order polynomial ground truth function. Subsequently, the performance of GPR with Matern and linear kernel in UAL procedure, as well as the performance of the corresponding random sampling strategy are assessed.

\begin{figure}
    \centering
    \includegraphics[width = 1\linewidth]{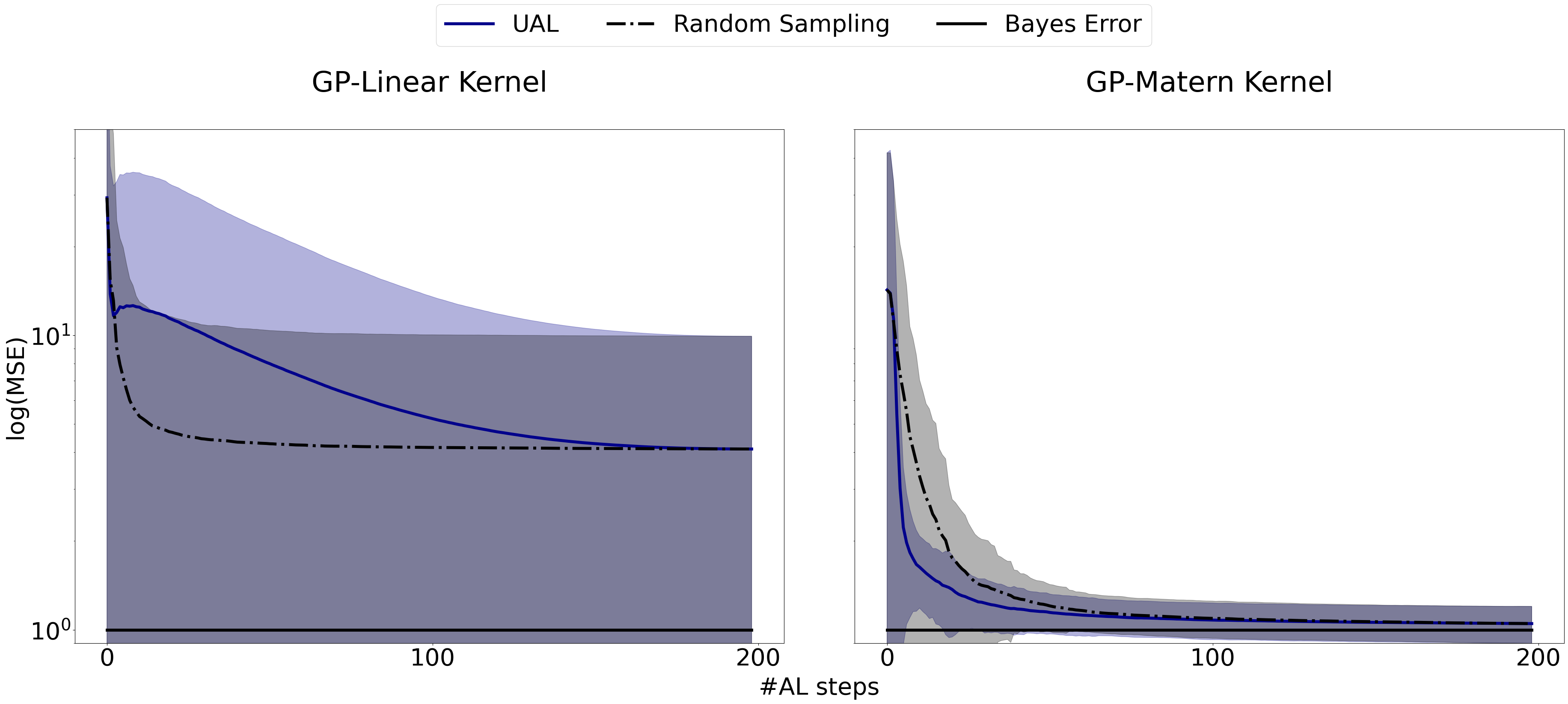}
    \caption{Performance of Matern and Linear kernels on GP regression task over the $3^{rd}$-order polynomial family. As the kernel becomes more complex, the ability of variance to faithfully capture the MSE increases. This is reflected in the improved UAL performance.}
    \label{fig:gp}
\end{figure}

Akin to observations in BPR, Figure~\ref{fig:gp} illustrates that in GPR with Matern kernel, UAL performs better than naive random sampling which is expected. However, with a linear kernel, UAL performs significantly worse than random sampling. 
We make the same explanation as BPR here, i.e., based on the analysis in Section \ref{UAL efficacy}, when the model's capacity equals or exceeds the ground truth, the predictive variance captures the actual MSE that leads to the better performance of UAL. Furthermore, the incapability of model class (linear kernel) with low capacity results in an inaccurate estimation of MSE. This mismatch leads to a degraded UAL performance.

\begin{figure}[b]\vspace{-4mm}
    \centering
    \includegraphics[width=0.6\linewidth]{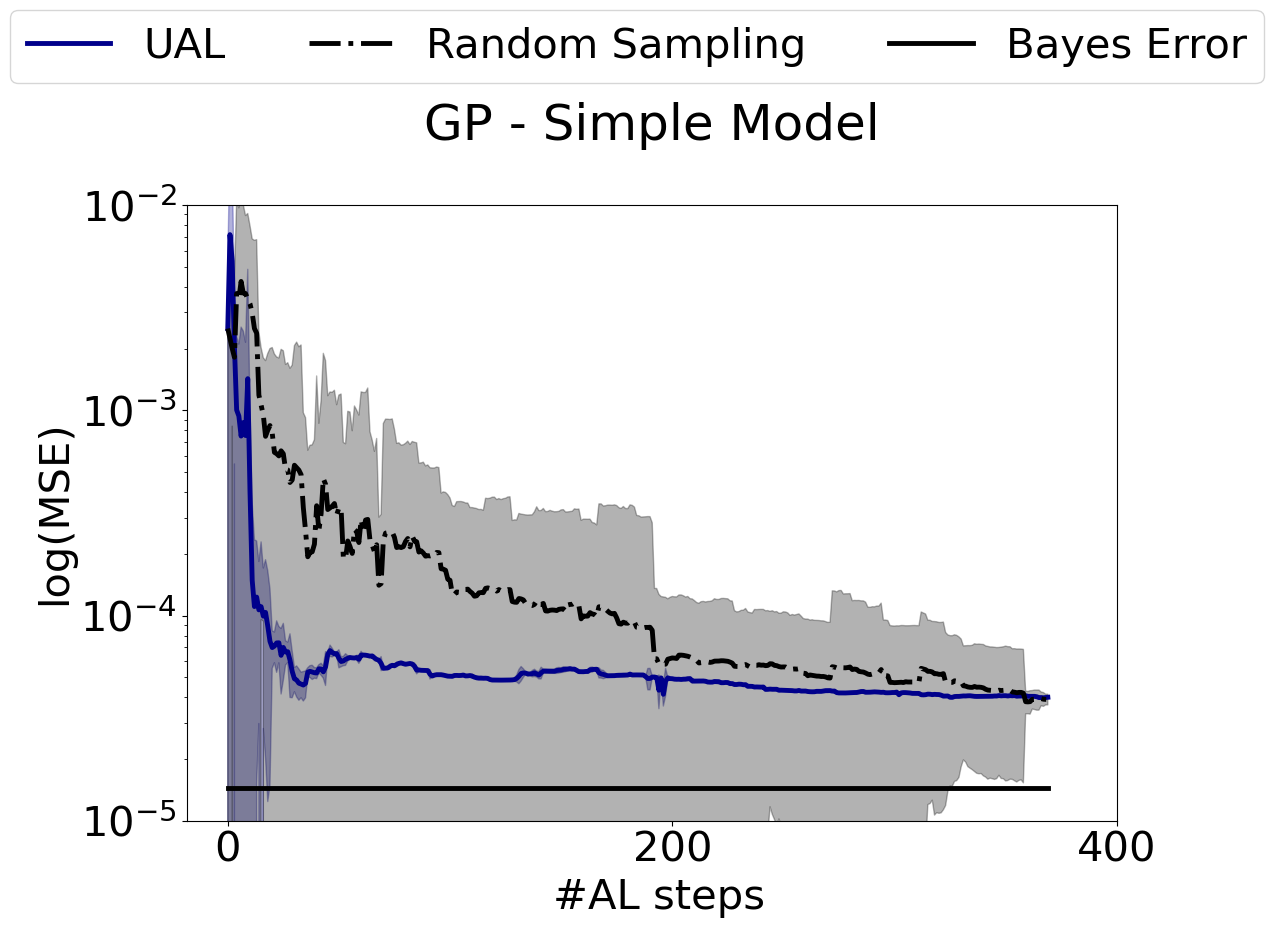}
    \caption{UAL performance using a simple model, GPR with a linear kernel, on the \emph{Facebook Metrics} dataset. The model is able to capture the underlying feature-response relationship and successfully directs the UAL process, resulting in superior performance compared to random sampling.}
    \label{fig:FB-dataset}
\end{figure}

\begin{figure}
    \centering
    \includegraphics[width=1\linewidth]{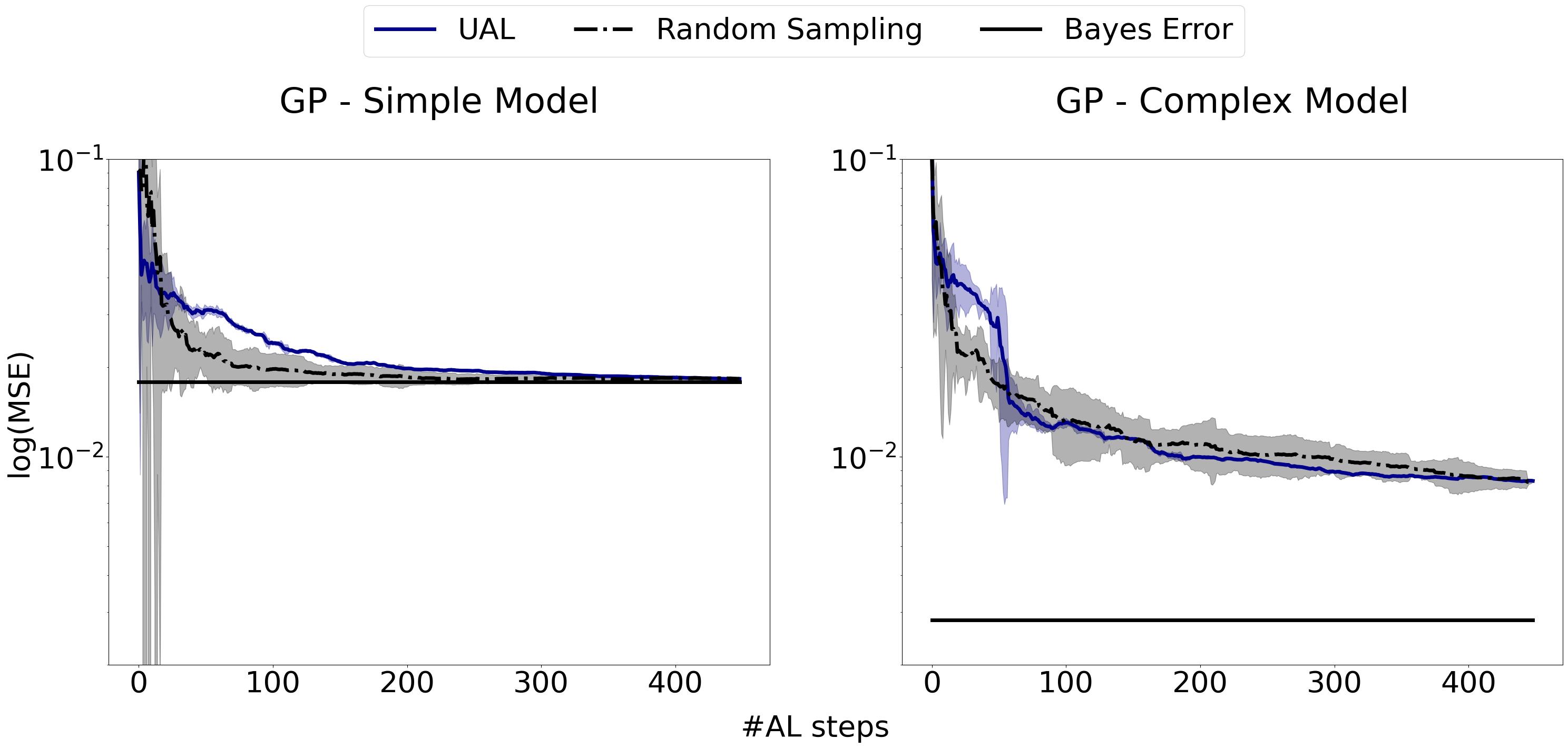}
    \caption{UAL performance using simple (left) and complex (right) models on the \emph{Concrete Compressive Strength} dataset. It is clear that UAL with the simple model is incapable of efficiently guiding the model learning procedure.}
    \label{fig:Conc-dataset}
\end{figure}

{In real-world case studies, we report the evaluated performance of GPR on the \emph{Facebook Metrics} and \emph{Concrete Compressive Strength} datasets. For \emph{Facebook Metrics}, due to the uncomplicated nature of the dataset, we only report the performance of the simple model in Figure \ref{fig:FB-dataset}. Clearly, owing to the simplicity of the regression task on this dataset, the simple model efficiently guides the UAL process and outperforms the naive random sampling strategy. 
On the other hand, for the more challenging regression task on the \emph{Concrete Compressive Strength} dataset, as demonstrated in Figure \ref{fig:Conc-dataset}, the variance-based acquisition function with the simple model fails to direct UAL process, leading to poor performance in comparison to random sampling. However, the variance-based acquisition function with the complex model successfully guides the UAL process and delivers a superior performance than naive random sampling strategy.}

Now that we have established the issues of variance-based UAL when the target function to learn is outside of the model class, a natural question is whether we are 
able to make UAL's performance independent of its underlying model class. In what follows, we provide two potential simple solutions and empirically demonstrate their effectiveness. 

For both suggested solutions, the learning procedure starts with one randomly selected pair of input and output from the target; then, using a $1^{st}$-order polynomial predictive model under the same settings as in previous experiments, the performance of UAL and random sampling is assessed. 

\subsection{UAL by {Direct} Estimation of MSE}

{A reasonable approach is to design the acquisition function of UAL directly approximating the true objective, MSE. Since we do not know the ground truth target in practice, one potential solution is to adopt 
a surrogate model to estimate MSE for unlabeled data points by training it using the observed data, as mentioned in Section~\ref{remedy}. To illustrate this, we contemplate utilizing a Gaussian Process to approximately estimate MSE, similar as in Kriging~\cite{cressie1990origins}.
At each stage, we label the data point with the highest error estimated by GP and add it to the labeled dataset, then update the predictive model and GP after dataset augmentation.}

Here we give an example of such a UAL method. The ground truth in this example is the same target function as in Section \ref{Introduction} which is a $2^{nd}$-order polynomial plus a $cosine$ function and white noise.

\begin{figure}
    \centering
    \includegraphics[width = 1.02\linewidth]{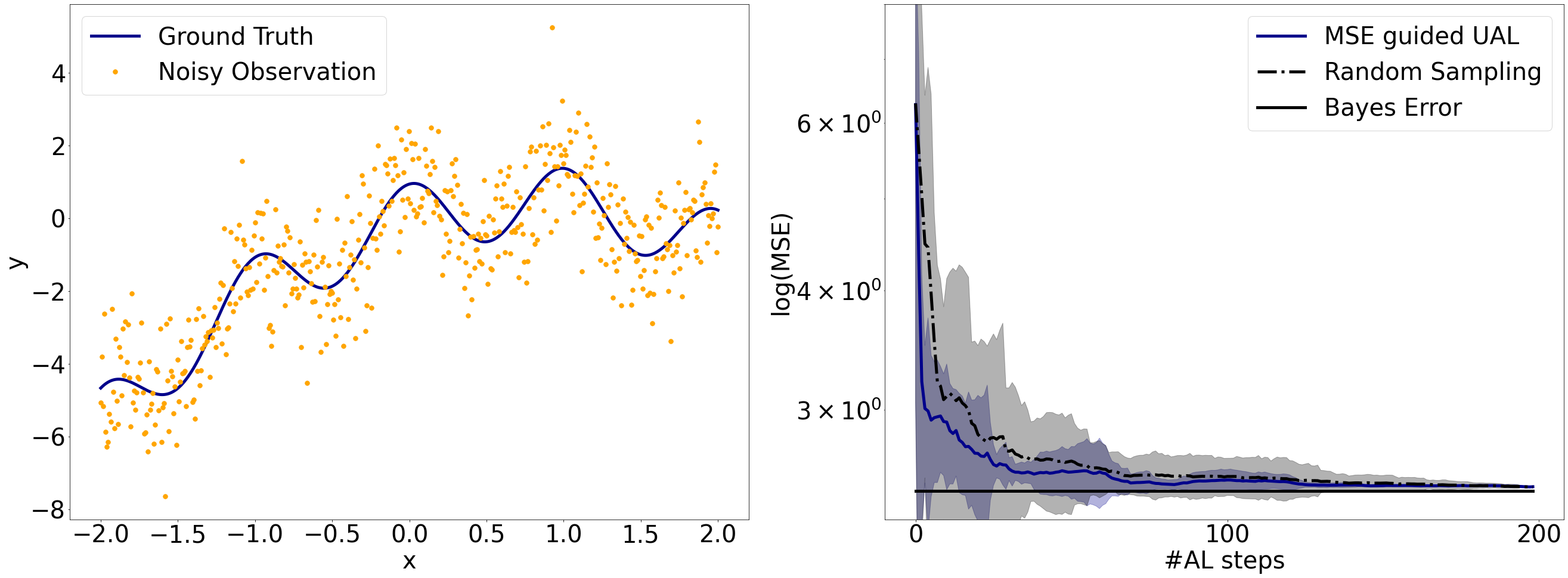}
    \caption{Example with a ground truth target more complex than the prediction model class. 
    In this example, the noisy data and ground truth function are the same as the motivating example in Section \ref{Introduction}. 
    The left plot shows the ground truth function and its corresponding noisy observed data. The right plot shows the comparison of MSE-guided UAL and random sampling to learn the linear predictor based on Bayesian polynomial regression.}
    \label{fig:eral-gp}
\end{figure}

Figure \ref{fig:eral-gp} illustrates the regression target and the regression performance of such a GP-approximated acquisition function. 
As shown, UAL guided by this approximate MSE estimation outperforms random sampling even when the model class does not cover the target. While here the challenge is how to better estimate MSE, which itself is an interesting future research direction. 

\subsection{UAL by Estimated Upper-bound of MSE}

Considering the rationale of the previous solution, we aim to predict the objective for each unseen sample. Also, as discussed previously, we only care about the relative value of the objective for each sample, meaning that only the pattern of the true objective matters and not the actual value. Following the second approach mentioned in Section~\ref{remedy}, an upper bound of MSE which captures its actual trend can be used as a functional acquisition function. 


In this part, to provide a solution based on the estimated upper bound, we will borrow the results provided in 
\cite{b713455c1b2c4ae28448b77823fe2a43}. In this work, by modeling the underlying function with a Gaussian Process and assuming {that} kernels are twice differentiable and translation-invariant, they have derived upper bounds for both the \textbf{Bias} and the \textbf{Variance} separately which is then used to bound the expected posterior loss. 
The only requirements for estimating the upper bounds are specifying the upper bounds on the gradient and the ground truth's function variance as prior knowledge. 

In the same manner, as all our experiments, we label the data point with the highest acquisition value (i.e., the highest estimated MSE upper bound), which will be added to the labeled dataset at each stage. Then we update the predictive model using the new augmented labeled dataset.

\begin{figure}
    \centering
    \includegraphics[width = 1.02\linewidth]{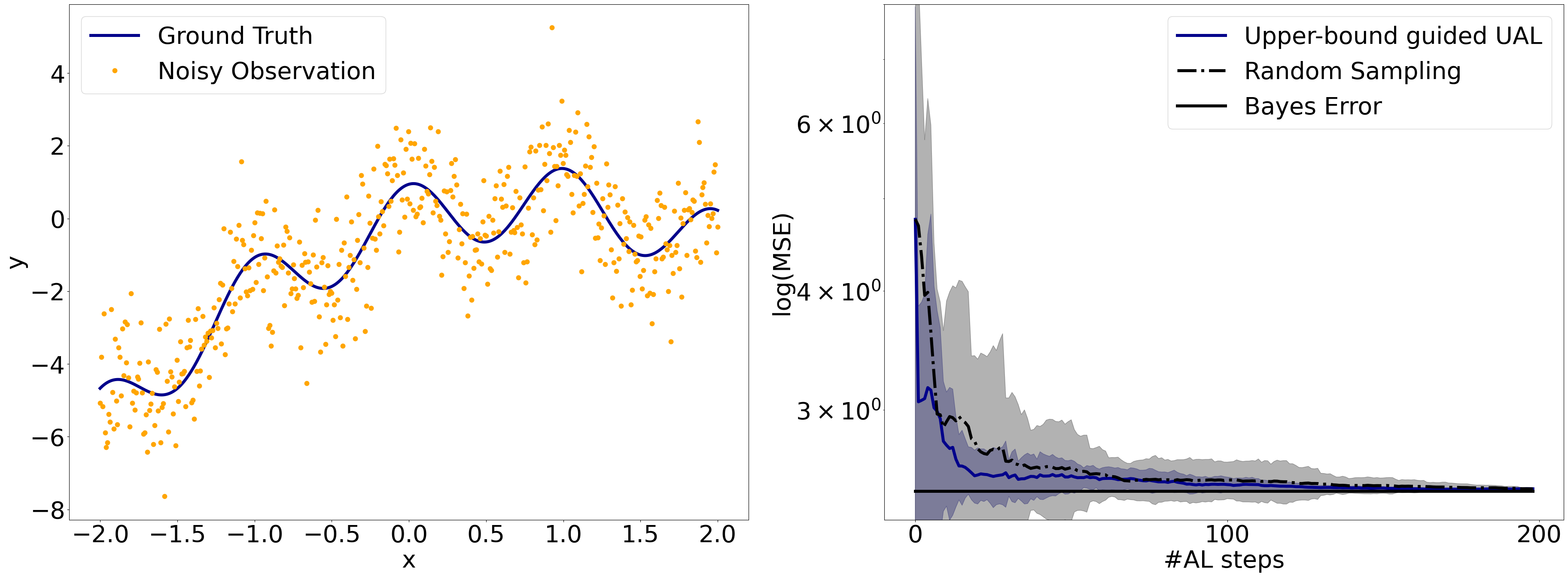}
    \caption{Example with the same target as the motivating example in Section \ref{Introduction} with a more complex ground truth than the prediction model class. 
    The left plot shows the ground truth function and its corresponding noisy observed data. The right plot shows the comparison of the UAL guided by the MSE's upper bound and random sampling to learn the linear predictor based on Bayesian polynomial regression.}
    \label{fig:ubal}
\end{figure}

With the same ground truth used in the previous solution, we illustrate the performance of the aforementioned solution. The regression 
efficacy 
is displayed in Figure~\ref{fig:ubal} (right-hand-side plot) alongside the ground truth (left-hand-side plot). It is clear that by using the MSE upper-bound as an acquisition function, UAL is potentially able to outperform random sampling even with a model class incapable of covering the target function. However, the tightness and more importantly how precise the upper bound is (concerning the true pattern of MSE), can improve UAL's performance and reliability. 

\section{Conclusions}\label{Conclusion}

The performance of UAL, one of the most common pool-based BAL strategies, heavily depends on the adopted acquisition function guiding sample selection. When the estimated uncertainty is used to define the acquisition function, the faithfulness of capturing model prediction performance affects the UAL efficacy.
We embark on a comprehensive investigation that delves into the analysis of UAL efficacy and the potential mismatch between the prediction model class complexities from the ground truth target. By analyzing the bias-variance decomposition of prediction error, we showed when the bias due to potential model mismatch dominates the prediction error, 
UAL may fail, performing worse than training based on random sampling. 
We conclude that for UAL to perform well, the choice of acquisition function is critical. When the prediction model class complexity aligns with or exceeds the intrinsic complexity of the target (i.e., ground truth model), prediction variance-guided UAL can perform better than random sampling. Otherwise, better acquisition functions 
dependent on the model's prediction performance may be needed for efficient UAL.

We have performed comprehensive empirical evaluations for both BPR and GPR, validating our theoretical analysis. By providing evaluations on two real-world datasets, we further demonstrated the reliability of our conclusions.
Furthermore, we have laid out that the estimation of the true objective might be a potential remedy for UAL shortcomings in case of model mismatch. Specifically, we provided a naive error-based acquisition function estimation strategy via kriging, which has the potential to achieve efficient UAL under model mismatch. Additionally, by taking advantage of the results in~\cite{b713455c1b2c4ae28448b77823fe2a43}, we presented results of another error-dependent acquisition function that estimates the objective's upper bound and shows promising UAL performance in such scenarios.
Theoretical bound analysis on the quantitative relationships between model mismatch and UAL efficacy as well as new acquisition functions based on reliable error estimation for UAL are potential future research directions to pursue.

\appendix

\section*{\textbf{MSE derivation}}
Here we give the detailed derivation for the MSE utilized in Section \ref{UAL efficacy}.

Defining $f:\mathcal{X}\rightarrow \mathbb{R}$ as the ground truth function, $y=f + \epsilon$ the noisy observation, and $\hat{f}_\theta$ as the model's prediction, where $f(\mathbf{x}) = \langle{\phi(\mathbf{x}},l),\mathbf{w}\rangle \in \mathbb{R}$ and $\hat{f}_\theta(x) = \langle{\phi(\mathbf{x}},p),\theta\rangle \in \mathbb{R}$, with $\epsilon \sim N(0,\sigma^2)$, $\phi(\mathbf{x},l) = \phi^l \in \mathbb{R}^{l}$, and $\phi(\mathbf{x},p) = \phi^p \in \mathbb{R}^{p}$, 
\[
\begin{aligned}
\mathbf{w} \sim N(\mu, \Sigma)\in \mathbb{R}^{(l+1)} &\quad  \mu \in \mathbb{R}^{(l+1)}, \quad \Sigma \in \mathbb{R}^{(l+1)\times (l+1)} \\
\theta \sim N(\hat{\mu}, \hat{\Sigma})\in \mathbb{R}^{(p+1)} &\quad \hat{\mu} \in \mathbb{R}^{(p+1)}, \quad \hat{\Sigma} \in \mathbb{R}^{(p+1)\times (p+1)} \\
P(\theta|\hat{\mathbf{\Phi}},\mathbf{y},\sigma^2)  \propto  N(\theta|\hat{\mu}, &\hat{\Sigma})N(y|\hat{\mathbf{\Phi}}\theta, \sigma^2)
    = N(\theta|\hat{\mu}_p, \hat{\Sigma}_p) \\
    \hat{\mu}_p = \hat{\Sigma}_p\hat{\Sigma}^{-1}\hat{\mu} + \frac{1}{\sigma^2}\hat{\Sigma}_p&\hat{\mathbf{\Phi}}^\top \mathbf{y},\quad
    \hat{\Sigma}_p^{-1} = \hat{\Sigma}^{-1} + \frac{1}{\sigma^2}\hat{\mathbf{\Phi}}^\top \hat{\mathbf{\Phi}}\\
    \pi^*(\hat{y}|\mathbf{x}) = \int N(\hat{y}&|\langle\phi^p,\theta\rangle, \sigma^{2})N(\theta|\hat{\mu}_p,\hat{\Sigma}_p)d\theta \\
    =& N(\hat{y}|\langle\phi^p,\hat{\mu}_p\rangle, \sigma_p^{2}(\mathbf{x}))\\
    \sigma_p^{2}(\mathbf{x}) =& \sigma^2 + (\phi^p)^\top \hat{\Sigma}_p (\phi^p) \\   
\end{aligned}
\]

Note that as mentioned in Section \ref{UAL efficacy}, we focus on $\text{MSE}$ of predicted output from the ground truth.

Using the above formulation, one can write the MSE as follows:
\begin{equation}\label{eq:mse_app}
\begin{aligned}
\text{MSE} 
    &= E_{P(Y|\mathbf{X})}[E_{P^*(\theta|\mathbf{D})}[ (f({\mathbf{x}})-\hat{f}_\theta({\mathbf{x}}))^2 ]]\\
    &= E_{P(Y|\mathbf{X})}[E_{P^*(\theta|\mathbf{D})}[(f(\mathbf{x})-\langle\phi^p,\hat{\mu}_p\rangle)^2]] \\
    &\quad+ E_{P(Y|\mathbf{X})}[E_{P^*(\theta|\mathbf{D})}[(\langle\phi^p,\hat{\mu}_p\rangle-\langle\phi^p,\theta\rangle)^2]] \\
    &= E_{P(Y|\mathbf{X})}[(f(\mathbf{x})-\langle\phi^p,\hat{\mu}_p\rangle)^2] + (\phi^p)^\top \hat{\Sigma}_p(\phi^p) \\
    &= E_{P(\mathbf{w})}[E_{P(Y|\mathbf{w},\mathbf{X})}[(f(\mathbf{x})-\langle\phi^p,\hat{\mu}_p\rangle)^2]]  + (\phi^p)^\top \hat{\Sigma}_p(\phi^p)\\
    &= E_{P(\mathbf{w})}[E_{P(Y|\mathbf{w},\mathbf{X})}[(\langle\phi^l,\mathbf{w}\rangle-\langle\phi^p,\hat{\mu}_p\rangle)^2]] \\
    &\quad+ (\phi^p)^\top \hat{\Sigma}_p(\phi^p)\\
    &= E_{P(\mathbf{w})}[E_{P(Y|\mathbf{w},\mathbf{X})}[\langle\phi^l,\mathbf{w}\rangle^2 - 2\langle\phi^l,\mathbf{w}\rangle\langle\phi^p,\hat{\mu}_p\rangle  \\
    &\quad+ \langle\phi^p,\hat{\mu}_p\rangle^2]] + (\phi^p)^\top \hat{\Sigma}_p(\phi^p) \\
    &= E_{P(\mathbf{w})}[\langle\phi^l,\mathbf{w}\rangle^2 \\
    &- 2\langle\phi^l,\mathbf{w}\rangle\langle\phi^p,\hat{\Sigma}_p\hat{\Sigma}^{-1}\hat{\mu} + \frac{1}{\sigma^2}\hat{\Sigma}_p\hat{\mathbf{\Phi}}^\top \mathbf{\Phi}\mathbf{w}\rangle \\
    &\quad+ \langle\phi^p,\hat{\Sigma}_p\hat{\Sigma}^{-1}\hat{\mu}\rangle^2 + \frac{2}{\sigma^2}\langle\phi^p, \hat{\Sigma}_p\hat{\Sigma}\hat{\mu}\rangle\langle\phi^p,\hat{\Sigma}_p\hat{\mathbf{\Phi}}^\top \mathbf{\Phi}\mathbf{w}\rangle \\
    &\quad+ \frac{1}{\sigma^4}\langle\phi^l,\hat{\Sigma}_p\hat{\mathbf{\Phi}}^\top \mathbf{\Phi}\mathbf{w}\rangle^2 + \frac{1}{\sigma^2}\langle\phi^p, \hat{\Sigma}_p\hat{\mathbf{\Phi}}^\top \rangle^2] \\
    &\quad+ (\phi^p)^\top \hat{\Sigma}_p(\phi^p) \\ 
    &= \langle\phi^l,\mu\rangle^2 + (\phi^l)^\top \Sigma(\phi^l) - 2\langle\phi^l,\mu\rangle\langle\phi^p,\hat{\Sigma}_p\hat{\Sigma}^{-1}\hat{\mu}\rangle \\
    &\quad- \frac{2}{\sigma^2}\langle\phi^l,\mu\rangle\langle\phi^p,\hat{\Sigma}_p\hat{\mathbf{\Phi}}^\top \mathbf{\Phi}\mu\rangle - \frac{2}{\sigma^2}(\phi^l)^\top \Sigma\mathbf{\Phi}^\top \hat{\mathbf{\Phi}}\hat{\Sigma}_p(\phi^l)\\
    &\quad+ \langle\phi^p,\hat{\Sigma}_p\hat{\Sigma}^{-1}\hat{\mu}\rangle^2 \quad+ \frac{2}{\sigma^2}\langle\phi^p, \hat{\Sigma}_p\hat{\Sigma}^{-1}\hat{\mu}\rangle\langle\phi^p,\hat{\Sigma}_p\hat{\mathbf{\Phi}}^\top \mathbf{\Phi}\mu\rangle \\
    &\quad+ \frac{1}{\sigma^4}\langle\phi^p,\hat{\Sigma}_p\hat{\mathbf{\Phi}}^\top \mathbf{\Phi}\mu\rangle^2 +\frac{1}{\sigma^4}(\phi^p)^\top \hat{\Sigma}_p\hat{\mathbf{\Phi}}^\top \mathbf{\Phi}\Sigma\mathbf{\Phi}^\top \hat{\mathbf{\Phi}}\hat{\Sigma}_p(\phi^p)\\
    &\quad+ \frac{1}{\sigma^2}\langle\phi^p, \hat{\Sigma}_p\hat{\mathbf{\Phi}}^\top \rangle^2 + (\phi^p)^\top \hat{\Sigma}_p(\phi^p)
\end{aligned}
\end{equation}

In the following, the MSE simplifications for the Matched Model and Lower-Order Model scenarios are provided:

\subsection*{\textbf{Matched Model ($\mathbf{p=l}$)}}
In this case, $\phi^p = \phi^l,\quad \hat{\mu} = \mu, \quad \hat{\Sigma} = \Sigma, \text{ and } \hat{\mathbf{\Phi}} = \mathbf{\Phi}$. Also, we know that $\hat{\mathbf{\Phi}}^\top \hat{\mathbf{\Phi}} = \sigma^2 (\hat{\Sigma}_p^{-1} - \Sigma^{-1})$. By replacing $\mathbf{\Phi}^\top \mathbf{\Phi},\quad \hat{\mathbf{\Phi}}^\top \mathbf{\Phi}, \text{ and} \hat{\mathbf{\Phi}}^\top \hat{\mathbf{\Phi}}$ with their equivalent, we try to simplify the MSE. 

\[
\small
\begin{aligned}
    \normalsize\text{MSE} 
    &= (\phi^l)^\top (\mu\mu^\top )(\phi^l) + (\phi^l)^\top (\Sigma)(\phi^l) \\
    &\quad- 2(\phi^l)^\top \hat{\Sigma}_p\Sigma^{-1}\mu\mu^\top (\phi^l)- 2(\phi^l)^\top \mu\mu^\top (\phi^l) \\
    &\quad+ 2(\phi^l)^\top \hat{\Sigma}_p\Sigma^{-1}\mu\mu^\top (\phi^l) -2(\phi^l)^\top \Sigma(\phi^l) \\
    &\quad+ 2(\phi^l)^\top \hat{\Sigma}_p(\phi^l) + (\phi^l)^\top \hat{\Sigma}_p\Sigma^{-1}\mu\mu^\top \Sigma^{-1}\hat{\Sigma}_p(\phi^l) \\
    &\quad+ (\phi^l)^\top \hat{\Sigma}_p\Sigma^{-1}\mu\mu^\top (\phi^l) - (\phi^l)^\top \hat{\Sigma}_p\Sigma^{-1}\mu\mu^\top \Sigma^{-1}\hat{\Sigma}_p(\phi^l)\\
    &\quad+ (\phi^l)^\top \mu\mu^\top \Sigma^{-1}\hat{\Sigma}_p(\phi^l) - (\phi^l)^\top \hat{\Sigma}_p\Sigma^{-1}\mu\mu^\top \Sigma^{-1}\hat{\Sigma}_p(\phi^l)\\
    &\quad+(\phi^l)^\top \mu\mu^\top (\phi^l)-(\phi^l)^\top \mu\mu^\top \Sigma^{-1}\hat{\Sigma}_p(\phi^l)\\
    &\quad-(\phi^l)^\top \hat{\Sigma}_p\Sigma^{-1}\mu\mu^\top (\phi^l)+(\phi^l)^\top \hat{\Sigma}_p\Sigma^{-1}\mu\mu^\top \Sigma^{-1}\hat{\Sigma}_p(\phi^l)\\
    &\quad+ (\phi^l)^\top \Sigma(\phi^l) - 2(\phi^l)^\top \hat{\Sigma}_p(\phi^l) + (\phi^l)^\top \hat{\Sigma}_p\Sigma^{-1}\hat{\Sigma}_p(\phi^l) \\&
    \quad+ (\phi^l)^\top \hat{\Sigma}_p(\phi^l)- (\phi^l)^\top \hat{\Sigma}_p\Sigma^{-1}\hat{\Sigma}_p(\phi^l) + (\phi^l)^\top \hat{\Sigma}_p(\phi^l)\\
    &= 2(\phi^l)^\top \hat{\Sigma}_p(\phi^l) = 2(\phi^p)^\top \hat{\Sigma}_p(\phi^p)= 2(\sigma_p(\mathbf{x})-\sigma^2)
    \end{aligned}
\]

\subsection*{\textbf{Lower-Order Model ($\mathbf{p<l}$)}}

Define\[
\begin{aligned}
    \mathbf{\Phi} =& [\tilde{\mathbf{\Phi}}_c \quad \hat{\mathbf{\Phi}}] \in \mathbb{R}^{n\times (l+1)}, \qquad
    \mu = \begin{bmatrix}
    \tilde{\mu}_{c} \\  \tilde{\mu}
    \end{bmatrix} \in \mathbb{R}^{(l+1)} \\
    \phi^l =& \begin{bmatrix}
        \tilde{\phi}_c \\ \phi^p 
    \end{bmatrix} \in \mathbb{R}^{(l+1)} ,\qquad \quad
    \Sigma = \begin{bmatrix}
        \tilde{\Sigma}_{c} \quad \Sigma_{12}\\ \\
        \Sigma_{12}^\top  \quad \tilde{\Sigma}
    \end{bmatrix} \in \mathbb{R}^{(l+1)\times (l+1)} 
\end{aligned}
\]
with $\tilde{\phi}_c = [x^{p+1}, \dots, x^{l}]$, where 
\[
\begin{aligned}
     \tilde{\mathbf{\Phi}}_c \in \mathbb{R}^{n\times(l-p)}, &\qquad \hat{\mathbf{\Phi}} \in \mathbb{R}^{n\times (p+1)}, \\
     \tilde{\mu}_{c} \in \mathbb{R}^{(l-p)}, &\qquad \tilde{\mu} \in \mathbb{R}^{ (p+1)} \\
     \tilde{\phi}_c \in \mathbb{R}^{(l-p)}, &\qquad \phi^{p} \in \mathbb{R}^{(p+1)}\\
    \tilde{\Sigma}_c \in \mathbb{R}^{(l-p)\times (l-p)},  \Sigma_{12} &\in \mathbb{R}^{(l-p)\times (p+1)},  \tilde{\Sigma} \in \mathbb{R}^{(p+1)\times (p+1)}
    \\
\end{aligned}
\]
Assuming $\tilde{\Sigma}=\hat{\Sigma}$ and $\tilde{\mu}=\hat{\mu}$ similar to Matched Model case, 
each term of MSE (Eq. \ref{eq:mse_app}) can be further derived:

\begin{itemize}
    \item {$1^{st} + 2^{nd}$  term:} 
    \[
    \begin{aligned}
    (\tilde{\phi}_c)^\top (\tilde{\Sigma}_{c} + \tilde{\mu}_{c}\tilde{\mu}_{c}^\top )(\tilde{\phi}_c) +& 2 (\tilde{\phi}_c)^\top (\Sigma_{12}+\tilde{\mu}_{c}\hat{\mu}^\top )(\phi^p) \\
     + (\phi^p)^\top (\hat{\Sigma}+&\hat{\mu}\hat{\mu}^\top )(\phi^p)       
    \end{aligned}
    \]
    \item {$3^{rd}$ term:}
    \[    (\phi^p)^\top \hat{\Sigma}_p\hat{\Sigma}^{-1}\hat{\mu}\tilde{\mu}_{c}^\top (\tilde{\phi}_c) +(\phi^p)^\top \hat{\Sigma}_p\hat{\Sigma}^{-1}\hat{\mu}\hat{\mu}^\top (\phi^p)
    \]
    \item {$4^{th}+5^{th}$ term:}
    \[
    \begin{aligned}
    (\phi^p)^\top \hat{\Sigma}_p\hat{\mathbf{\Phi}}^\top \tilde{\mathbf{\Phi}}_c(\tilde{\Sigma}_{c}& + \tilde{\mu}_{c}\tilde{\mu}_{c}^\top )(\tilde{\phi}_c) \\
    + (\phi^p)^\top \hat{\Sigma}_p\hat{\mathbf{\Phi}}^\top \tilde{\mathbf{\Phi}}_c({\Sigma}_{12}& + \tilde{\mu}_{c}\hat{\mu}^\top )(\phi^p) \\
    + (\phi^p)^\top \hat{\Sigma}_p\hat{\mathbf{\Phi}}^\top \hat{\mathbf{\Phi}}({\Sigma}_{12}^\top&  + \hat{\mu}\tilde{\mu}_{c}^\top )(\tilde{\phi}_c) \\
    +(\phi^p)^\top \hat{\Sigma}_p\hat{\mathbf{\Phi}}^\top \hat{\mathbf{\Phi}}(\hat{\Sigma}& + \hat{\mu}\hat{\mu}^\top )(\phi^p)
    \end{aligned}
    \]

    \item {$6^{th}$ term:} The same as in Eq. \eqref{eq:mse_app}
    \item {$7^{th}$ term:} 
    \[ 
    \begin{aligned} &\quad(\phi^p)^\top \hat{\Sigma}_p\hat{\Sigma}^{-1}\hat{\mu}\tilde{\mu}_{c}^\top \tilde{\mathbf{\Phi}}_c^\top \hat{\mathbf{\Phi}}\hat{\Sigma}_p(\phi^p)\\
    &+ (\phi^p)^\top \hat{\Sigma}_p\hat{\Sigma}^{-1}\hat{\mu}\hat{\mu}^\top \hat{\mathbf{\Phi}}^\top \hat{\mathbf{\Phi}}\hat{\Sigma}_p(\phi^p) 
    \end{aligned}
    \]
    \item {$8^{th}+9^{th}$ term:} 
    \[
   \begin{aligned} (\phi^p)^\top \hat{\Sigma}_p\hat{\mathbf{\Phi}}^\top \tilde{\mathbf{\Phi}}_c(\tilde{\Sigma}_{c}&+\tilde{\mu}_{c}\tilde{\mu}_{c}^\top )\tilde{\mathbf{\Phi}}_c^\top \hat{\mathbf{\Phi}}\hat{\Sigma}_p(\phi^p) \\
   +2(\phi^p)^\top \hat{\Sigma}_p\hat{\mathbf{\Phi}}^\top \tilde{\mathbf{\Phi}}_c({\Sigma}_{12}&+\tilde{\mu}_{c}\hat{\mu}^\top )\tilde{\mathbf{\Phi}}_c^\top \hat{\mathbf{\Phi}}\hat{\Sigma}_p(\phi^p) \\
   + (\phi^p)^\top \hat{\Sigma}_p\hat{\mathbf{\Phi}}^\top \hat{\mathbf{\Phi}}(\hat{\Sigma}&+\hat{\mu}\hat{\mu}^\top )\tilde{\mathbf{\Phi}}_c^\top \hat{\mathbf{\Phi}}\hat{\Sigma}_p(\phi^p)
    \end{aligned}
    \]

    \item {$10^{th}$ and $11^{th}$ terms:} the same as in Eq. \eqref{eq:mse_app}. 
\end{itemize}

Notice that $\hat{\mathbf{\Phi}}^\top \hat{\mathbf{\Phi}} = \sigma^2(\hat{\Sigma}_p^{-1} - \hat{\Sigma}^{-1})$.

After rewriting Eq. \eqref{eq:mse_app}, the simplified MSE is as follows: 
\[
\small
\begin{aligned}
\text{MSE} = (\tilde{\phi}_c)^\top (\tilde{\Sigma}_{c}+\tilde{\mu}_{c}\tilde{\mu}_{c}^\top )(\tilde{\phi}_c) -&\frac{2}{\sigma^2} (\phi^p)^\top \hat{\Sigma}_p\hat{\mathbf{\Phi}}^\top \tilde{\mathbf{\Phi}}_c(\tilde{\Sigma}_{c}+\tilde{\mu}_{c}\tilde{\mu}_{c}^\top )(\tilde{\phi}_c) \\
+2(\phi^p)^\top \hat{\Sigma}_p&\hat{\Sigma}^{-1}\Sigma_{12}^\top (\tilde{\phi}_c)\\
+\frac{1}{\sigma^4} (\phi^p)^\top \hat{\Sigma}_p\hat{\mathbf{\Phi}}^\top \tilde{\mathbf{\Phi}}_c(\tilde{\Sigma}_{c} &+ \tilde{\mu}_{c}\tilde{\mu}_{c}^\top )\tilde{\mathbf{\Phi}}_c^\top \hat{\mathbf{\Phi}}\hat{\Sigma}_p(\phi^p) \\
- \frac{2}{\sigma^2} (\phi^p)^\top \hat{\Sigma}_p \hat{\mathbf{\Phi}}^\top \tilde{\mathbf{\Phi}}_c\Sigma_{12}&\hat{\Sigma}^{-1}\hat{\Sigma}_p(\phi^p)  + 2(\phi^p)^\top \hat{\Sigma}_p(\phi^p) \\
\end{aligned}
\]

\section*{\textbf{Proof of Proposition~\ref{prop:bias}}}
\begin{proposition}
    $(E_{P^*(\theta|\mathbf{D_L})}[\hat{f}_{\theta}]-  E_{P(Y|\mathbf{X})}[Y])^2 < \varepsilon^2 C^2$ if $E_{P(Y|\mathbf{X})}[|Y|]<C$ and $\big|\frac{\pi^*}{P(Y|\mathbf{X})}-1\big|<\varepsilon$, where $C > 0$ and $\varepsilon > 0$ are constants.
\end{proposition}
\begin{proof}
We first rewrite the \textbf{Bias} term $(E_{P^*(\theta|\mathbf{D_L})}[\hat{f}_{\theta}]-  E_{P(Y|\mathbf{X})}[Y])^2$ as
\[
\begin{aligned}
    \textbf{Bias} &= (E_{P^*(\theta|\mathbf{D_L})}[\hat{f}_{\theta}] - E_{P(Y|\mathbf{X})}[Y])^2 \\
    &=  (E_{P^*(\theta|\mathbf{D_L})}[E_{P(\hat{y}|\theta, \mathbf{X})}[\hat{y}]] - E_{P(Y|\mathbf{X})}[Y])^2\\
    &= (\int P^*(\theta|\mathbf{D_L}) \int \hat{y}P(\hat{y}|\theta, \mathbf{X})d\hat{y}  d\theta  - \int P(Y|\mathbf{X})YdY )^2\\
    &=(\int \hat{y} \pi^* d\hat{y}    - \int P(Y|\mathbf{X})YdY )^2 \\
    &=(\int Y \pi^* dY   - \int P(Y|\mathbf{X})YdY )^2 \\
    &=(\int Y (\pi^* - P(Y|\mathbf{X}))dY )^2\\
    &= (\int_{0}^{\infty} Y (\pi^* - P(Y|\mathbf{X}))dY )^2 \\
    &\quad+ (\int_{-\infty}^{0} Y (\pi^* - P(Y|\mathbf{X}))dY )^2 \\
    &\quad+ 2\Bigg(\int_{0}^{\infty} Y (\pi^* - P(Y|\mathbf{X}))dY \Bigg)\\
    &\qquad\qquad\Bigg(\int_{-\infty}^{0} Y (\pi^* - P(Y|\mathbf{X}))dY \Bigg). \\
\end{aligned}
\]
With $|\frac{\pi^*}{P(Y|\mathbf{X})}-1|<\varepsilon$, or equivalently, $-\varepsilon<\frac{\pi^*}{P(Y|\mathbf{X})}-1<\varepsilon$,
\[
\begin{aligned}
 &   -\varepsilon P(Y|\mathbf{X})<{\pi^*}-{P(Y|\mathbf{X})}<\varepsilon P(Y|\mathbf{X}).
\end{aligned}
\]

So that $-\varepsilon\int_{0}^{\infty}YP(Y|\mathbf{X})dY\leq\int_{0}^{\infty} Y (\pi^* - P(Y|\mathbf{X}))dY\leq\varepsilon\int_{0}^{\infty}YP(Y|\mathbf{X})dY$, and $\varepsilon\int_{-\infty}^{0}YP(Y|\mathbf{X})dY\leq\int_{-\infty}^{0} Y (\pi^* - P(Y|\mathbf{X}))dY\leq-\varepsilon\int_{-\infty}^{0}YP(Y|\mathbf{X})dY$, i.e.
\[
\begin{aligned}
    &|\int_{0}^{\infty} Y (\pi^* - P(Y|\mathbf{X}))dY| \leq  \varepsilon\int_{0}^{\infty}YP(Y|\mathbf{X})dY ;\\
    &|\int_{-\infty}^{0} Y (\pi^* - P(Y|\mathbf{X}))dY| \leq - \varepsilon\int_{-\infty}^{0}YP(Y|\mathbf{X})dY.
\end{aligned}
\]
So
\[
\footnotesize
\begin{aligned}
    \textbf{Bias}&= (\int_{0}^{\infty} Y (\pi^* - P(Y|\mathbf{X}))dY )^2 + (\int_{-\infty}^{0} Y (\pi^* - P(Y|\mathbf{X}))dY )^2 \\
    &\quad+ 2(\int_{0}^{\infty} Y (\pi^* - P(Y|\mathbf{X}))dY )(\int_{-\infty}^{0} Y (\pi^* - P(Y|\mathbf{X}))dY )\\
    &\leq \varepsilon^2 \times \big[  (\int_{0}^{\infty}YP(Y|\mathbf{X})dY)^2 + (\int_{-\infty}^{0}YP(Y|\mathbf{X})dY)^2 \\
    &\quad-2(\int_{0}^{\infty}YP(Y|\mathbf{X})dY)(\int_{-\infty}^{0}YP(Y|\mathbf{X})dY)   \big] \\ 
    &= \varepsilon^2 \times \big[  (\int_{0}^{\infty}|Y|P(Y|\mathbf{X})dY)^2 + (\int_{-\infty}^{0}|Y|P(Y|\mathbf{X})dY)^2 \\
    &\quad+2(\int_{0}^{\infty}|Y|P(Y|\mathbf{X})dY)(\int_{-\infty}^{0}|Y|P(Y|\mathbf{X})dY)   \big] \\ 
    & = \varepsilon^2 E^2_{P(Y|\mathbf{X})}[|Y|],
\end{aligned}
\]
with $E_{P(Y|\mathbf{X})}[|Y|]\leq C$, 
\[
\textbf{Bias} \leq \varepsilon^2 (E_{P(Y|\mathbf{X})}[|Y|])^2 \leq \varepsilon^2 C^2.
\]
\end{proof}

\section*{Acknowledgments}
\noindent This work has been supported in part by the U.S. National Science Foundation (NSF) grant IIS-2212419; and by the U.S. Department of Energy (DOE) Office of Science, Advanced Scientific Computing Research (ASCR) M2DT Mathematical Multifaceted Integrated Capability Center (MMICC) under Award B\&R\# KJ0401010/FWP\# CC130, program manager W. Spotz. Portions of this research were conducted with the advanced computing resources provided by Texas A\&M High Performance Research Computing.

\bibliography{refs}
\bibliographystyle{IEEEtran}

\end{document}